\documentstyle{article}
\begin{document}
\bibliographystyle{plain}
\newtheorem{theorem}{Theorem}
\newtheorem{corollary}{Corollary}
\newtheorem{lemma}{Lemma}
\newtheorem{exercise}{Exercise}
\newtheorem{claim}{Claim}
\newtheorem{definition}{Definition}
\newenvironment{notation}{\noindent\bf Notation:\em\penalty1000}{}
\newcommand{\blackslug}{\mbox{\hskip 1pt \vrule width 4pt height 8pt 
depth 1.5pt \hskip 1pt}}
\newcommand{\QED}{\quad\blackslug\lower 8.5pt\null\par}
\newcommand{\proof}{\par\penalty-1000\vskip .5 pt\noindent{\bf Proof\/: }}
\newcommand{\ru}{\rule[-0.4mm]{.1mm}{3mm}}
\newcommand{\nni}{\ru\hspace{-3.5pt}}
\newcommand{\sni}{\ru\hspace{-1pt}}
\newcommand{\pre}{\hspace{0.28em}}
\newcommand{\post}{\hspace{0.1em}}
\newcommand{\NIm}{\pre\nni\sim}
\newcommand{\NI}{\mbox{ $\nni\sim$ }}
\newcommand{\notNIm}{\pre\nni\not\sim}
\newcommand{\notNI}{\mbox{ $\nni\not\sim$ }}
\newcommand{\NIVm}{\pre\nni\sim_V}
\newcommand{\NIV}{\mbox{ $\nni\sim_V$ }}
\newcommand{\notNIVm}{\pre\sni{\not\sim}_V\post}
\newcommand{\notNIV}{\mbox{ $\sni{\not\sim}_V$ }}
\newcommand{\NIWm}{\pre\nni\sim_W}
\newcommand{\NIW}{\mbox{ $\nni\sim_W$ }}
\newcommand{\NIWp}{\mbox{ $\nni\sim_{W'}$ }}
\newcommand{\notNIWm}{\pre\sni{\not\sim}_W\post}
\newcommand{\notNIW}{\mbox{ $\sni{\not\sim}_W$ }}
\newcommand{\eem}{\hspace{0.8mm}\rule[-1mm]{.1mm}{4mm}\hspace{-4pt}}
\newcommand{\EM}{\eem\equiv}
\newcommand{\notEM}{\eem\not\equiv}
\newcommand{\R}{\cal R}
\newcommand{\notR}{\not {\hspace{-1.5mm}{\cal R}}}
\newcommand{\bK}{{\bf K}}
\newcommand{\bP}{{\bf P}}
\newcommand{\ga}{\mbox{$\alpha$}}
\newcommand{\gb}{\mbox{$\beta$}}
\newcommand{\gc}{\mbox{$\gamma$}}
\newcommand{\gd}{\mbox{$\delta$}}
\newcommand{\cA}{\mbox{${\cal A}$}}
\newcommand{\cB}{\mbox{${\cal B}$}}
\newcommand{\cK}{\mbox{${\cal K}$}}
\newcommand{\cM}{\mbox{${\cal M}$}}
\newcommand{\cU}{\mbox{${\cal U}$}}
\newcommand{\ab}{\mbox{\ga \NI \gb}}
\newcommand{\ra}{\rightarrow}
\newcommand{\Ra}{\Rightarrow}
\newcommand{\eqdef}{\stackrel{\rm def}{=}}
\parskip=.080in     
\hyphenation{mono-tony Mono-tony mono-tonic}
\title{Nonmonotonic Reasoning, Preferential Models 
and Cumulative Logics
\thanks{This work was 
partially supported by the Jean and Helene Alfassa fund for 
research in Artificial Intelligence}}
\author{Sarit Kraus\thanks{Institute for Advanced Computer Studies 
and Department of Computer Science, 
University of Maryland, College Park, MD 20742 U.S.A. 
This work was done
while the author was at Hebrew University.},
\ Daniel Lehmann\thanks{Department of Computer Science, 
Hebrew University, Jerusalem 91904 (Israel)}
and Menachem Magidor\thanks{Department of Mathematics, 
Hebrew University, Jerusalem 91904 (Israel)}
}
\date{}
\maketitle

\begin{abstract}
Many systems that exhibit nonmonotonic behavior have been
described and studied already in the literature.
The general notion of
nonmonotonic reasoning, though, has almost always 
been described only negatively,
by the property it does not enjoy, i.e. monotonicity. 
We study here general patterns
of nonmonotonic reasoning and try to isolate properties that could
help us map the field of nonmonotonic reasoning by 
reference to positive properties.
We concentrate on a number of families of nonmonotonic consequence
relations, defined in the style of Gentzen~\cite{Gent:69}.
Both proof-theoretic and semantic points of view are developed in parallel.
The former point of view was pioneered by D.~Gabbay in \cite {Gabbay:85}, 
while the latter has been advocated by Y.~Shoham in \cite {Shoham:87}. 
Five such families are defined
and characterized by representation theorems, relating the two points of
view.
One of the families of interest, that of preferential relations, turns out
to have been studied by E.~Adams in \cite{Adams:75}. The {\em preferential}
models proposed here are a much stronger tool than
Adams' probabilistic semantics.
The basic language used in this paper is that of propositional
logic. The extension of our results to first order predicate calculi and
the study of the computational complexity of
the decision problems described in this paper 
will be treated in another paper. 
\end{abstract}

\section{Introduction}
\subsection{Nonmonotonic reasoning}
Nonmonotonic logic is the study of those ways of inferring additional 
information from given information that do not satisfy the monotonicity 
property satisfied by all methods based on classical (mathematical) logic.
In Mathematics, if a conclusion is warranted on the basis of 
certain premises, no additional premises
will ever invalidate the conclusion.

In everyday life, however, it seems clear that we, human beings, 
draw sensible conclusions from what we know and that, 
on the face of new information, 
we often have to take back previous conclusions, 
even when the new information
we gathered in no way made us want to take back our previous assumptions. 
For example, we may hold the assumption that most birds fly,
but that penguins are birds that do not fly and, learning
that Tweety is a bird, infer that it flies. Learning that Tweety is
a penguin, will in no way make us change our mind about the fact that
most birds fly and that penguins are birds that do not fly, 
or about the fact that Tweety is a bird. It should make
us abandon our conclusion about its flying capabilities, though.
It is most probable that intelligent automated systems will have 
to do the same kind of (nonmonotonic) inferences.

Many researchers have proposed systems that perform
such nonmonotonic inferences. The best known are probably: 
negation as failure \cite{Clark:78}, circumscription
\cite{McCarthy:80}, the modal system of \cite{McDer:80},
default logic \cite{Reiter:80}, 
autoepistemic logic \cite{Moo:84} and inheritance systems \cite{Tour:86}.
Each of those systems is worth studying by itself, 
but a general framework
in which those many examples could be compared and classified is missing.
We provide here a first attempt at such a general framework, concentrating
on properties that are or should be enjoyed by at least important 
families of nonmonotonic reasoning systems.
An up-to-date survey of the field of nonmonotonic reasoning may be found 
in \cite {Reiter:87}.

\subsection{Nonmonotonic consequence relations}
The idea that the best framework to study the deduction process is that of 
consequence relations dates back to A.~Tarski \cite{Tar:30}, 
\cite{Tar:30a} and \cite{Tar:35} (see \cite{Tar:56} 
for an English translation) and G.~Gentzen \cite{Gent:32} 
(see \cite{Gent:69}
for an English translation and related papers). 
For an up-to-date view on monotonic consequence relations, the reader may 
consult \cite{Avron:87}. 
Tarski studied the consequences of arbitrary sets
of formulas whereas Gentzen restricted himself to finite such sets.
In the presence of compactness, the difference between the two approaches
is small for monotonic consequence relations. For nonmonotonic relations,
many different notions of compactness come to mind, and the relation
between Tarski's infinitistic approach and Gentzen's finitistic approach
is much less clear. We develop here a finitistic approach in the style of 
Gentzen.
In \cite{Mak:89}, D.~Makinson developed, in parallel with and
independently from our effort, an infinitistic view of nonmonotonic
consequence relations. Later efforts in this direction, by M.~Freund and
D.~Lehmann~\cite{FLM:89}, 
have benefited from the results presented here.

D.~Gabbay \cite {Gabbay:85} was probably the first to suggest to focus the
study of nonmonotonic logics on their consequence relations. This is a
bold step to take since some of the nonmonotonic systems 
mentioned above were not meant to define a consequence relation, as was
soon noticed by D.~Israel in \cite{Isr:80}.
D.~Gabbay asked the question: what are the minimal conditions a
consequence relation should satisfy to represent a bona fide 
nonmonotonic {\em logic}?
He proposed three:
reflexivity (see equation~\ref{ru:ref} in section~\ref{subsec:cum}), 
cut (see equation~\ref{ru:cut})
and weak monotonicity (see equation~\ref{ru:caut}). 
Weak monotonicity has, since, been 
renamed cautious monotonicity by D.~Makinson \cite{Mak:89} and we shall
follow this last terminology, notwithstanding the fact that D.~Makinson 
has now 
opted for the term {\em cumulative monotony}. 
D.~Gabbay argued for his three conditions on 
proof-theoretic grounds but provided no semantics against which to check them.
He also assumed a $poor$ underlying language for propositions,
a language without classical propositional connectives.
In \cite{Mak:89}, D.~Makinson proposed a semantics for Gabbay's 
logic and proved a completeness result, for a {\em poor} language. 
His models have a definitely syntactic flavor, whereas the
models presented here seem more truly semantic and more easily suggest
rules of inference.

Independently, Y.~Shoham in \cite{Shoham:88} and \cite{Shoham:87}
proposed a general model theory for nonmonotonic inference.
He suggested models that may be described as
a set of worlds equipped with a {\em preference}
relation: the preference relation is a partial order 
and a world $v$ is preferable, in the eyes of the reasoner, to
some other world $w$ if he considers $v$ to be more $normal$ than $w$.
One would then, in the model, on the basis of a proposition $\alpha$,
conclude, defeasibly, that
a proposition $\beta$ is true 
if all worlds that satisfy $\alpha$ and are
{\em most normal} among worlds satisfying $\alpha$ also satisfy $\beta$.
Shoham claimed that adequate semantics could be given to 
known nonmonotonic
systems by using such a preference relation. 
He assumed a {\em rich} underlying
language for propositions, 
containing all classical propositional connectives.
The idea that nonmonotonic deduction should be modeled by some {\em
normality} relation between worlds is very natural and may be 
traced back to 
J.~McCarthy.
It appears also in relation with epistemic logic in \cite {HalpMoses:84}. 
One of the conclusions of this paper will be that none of the nonmonotonic
systems defined so far in the literature, except those based on conditional 
logic described in \cite{Del:87}, \cite{Del:88} and \cite{PearlGeff:88},
may represent all nonmonotonic inference systems that may be defined
by preferential models. The framework of preferential models, therefore,
has an expressive power that cannot be captured by negation as failure,
circumscription, default logic or autoepistemic logic. We do not claim that
all this expressive power is needed, 
but will claim that the systems mentioned
above lack expressive power.

The main point of this work, therefore, is to characterize the 
consequence relations that can be defined by models similar to Shoham's 
in terms of proof-theoretic properties.
To this end Gabbay's conditions 
have to be augmented. The class of models corresponding exactly to Gabbay's
conditions is also characterized.
The elucidation of the relations between proofs and models that is achieved
in this paper will allow for the design of decision procedures tuned to
different restrictions on the language of propositions or the knowledge
bases. Such decision procedures (or heuristics) could be the core of automated
engines of sensible inferences.
This paper will not propose any specific system of nonmonotonic reasoning.
Important steps towards such a system, taken after obtaining the
results reported here but before the final redaction of this paper, 
are reported in \cite{L:88}, \cite{LMTR:88} and \cite{Leh:89}.

At this point it could be useful to state the philosophy 
of this paper concerning
the relative importance of proof-theory and semantics. We consider, in this
paper, the axiomatic systems as the main object of interest (contrary to the
point of view expressed in \cite{Lewis:73} for example). 
The different families
of models described in this paper and that provide semantics to the axiomatic 
systems are not considered to be an ontological justification for our interest
in the formal systems, but only as a technical tool to study those systems and
in particular settle questions of interderivability and 
find efficient decision procedures.
Preliminary versions of the material contained in this paper appeared in
\cite{KL:88} and \cite{KLMTR:88}.

\subsection{Conditional logic}
In this subsection, the relation between our work and conditional logic will
be briefly surveyed. Since the link, we claim, is mainly at the level of the 
formal systems and not at the semantic level, the reader uninterested
in conditional logic may easily skip this subsection.

This work stems from a very different motivation than the vast body of work 
concerned with conditional logic and its semantics, (see in particular
\cite {Stal:68},\cite {Lewis:73} and \cite{Lewis:71}) 
which is surveyed in \cite{Nute:84}. Two main differences must be pointed at.
The first difference is
that conditional logic considers a binary intensional connective
that can be embedded inside other connectives and even itself,
whereas we consider a binary relation symbol that is part of 
the meta-language.
The second difference is that the semantics of the conditional 
implication of
conditional logic is essentially different from ours.
In conditional logic the formula 
\mbox{$\alpha > \beta$} is interpreted to mean {\em if $\alpha$ were (or was)
true and the situation were as close as possible, under
this hypothesis,
to what it really is, then $\beta$ would be true}.
For us \ab\ means that \ga\ is a good enough reason to believe \gb, or
that \gb\ is a plausible consequence of \ga.
The main difference is that conditional logic refers implicitly to the
{\em actual} state of the world whereas we do not. 
M.~Ginsberg's \cite{Gin:86}
proposal to harness conditional logic to nonmonotonic reasoning was clearly
set with the former semantics in mind, and that explains our disagreements
concerning the desirability of certain rules, e.g., the rule of
{\bf Rational Monotonicity}
(see equation (\ref{Rat:mon})).

One of the logical systems, {\bf P}, studied in this paper turns out to
be the flat (i.e. non-nested) fragment of a conditional logic
studied by J.~Burgess in 
\cite{Burgess:81} and by F.~Veltman in \cite{Velt:86}.
Because of their richer language, the semantics proposed in those 
papers are more complex than ours: a ternary relation of accessibility
between worlds is used in place of our binary preference relation. 
Moreover, the semantics of J.~Burgess are quite different from ours in some
other aspects; our semantics are
closer to F.~Veltman's (private communication from J.~van Benthem)
and to those studied by
J.~van Benthem in \cite{vBent:84}.
There are some 
connections between one of our completeness proofs and theirs, 
but the restricted 
language considered here simplifies the models and the proof a great deal.
Our completeness result cannot be derived from the completeness result of
\cite{Burgess:81} since the latter concerns a extended language and it is
not clear that a proof in the extended language may be translated in the 
restricted one.

This very fragment had been considered by E.~Adams in 
\cite {Adams:75} (see also \cite {Adams:66} for an earlier version and
motivation). E.~Adams' purpose was to propose probabilistic semantics
for indicative conditionals and not the study of nonmonotonic logics.
Recently J.~Pearl and H.~Geffner \cite {PearlGeff:88} have built upon 
E.~Adams' logics, our system {\bf P}, and his motivation in an effort to 
provide a system for nonmonotonic reasoning. For a gentle introduction,
see chapter 10 of \cite{Pearl:88}.
The semantics proposed here are not probabilistic. 
Probabilistic semantics that are equivalent with a restricted
family of models (ranked models) will be described elsewhere.
The preferential models
presented in this paper provide a much sharper
understanding of the system
{\bf P} than can obtained by Adams' methods.

\subsection{Plan of this paper}
This paper first describes the syntax proposed and compares it to more 
classical nonmonotonic systems.
Five logical systems and families of models are then presented
in turn and five soundness and completeness results are proven. 
The first system, {\bf C}, corresponds to D.~Gabbay's proposal.  
The second, stronger, system, {\bf CL}, includes a rule of inference 
that seems original, and corresponds to models that seem to be more
natural.
None of those systems above assumes, in any essential way 
the existence of the classical 
logical connectives, if one allows a finite set of formulas to appear on
the left of our symbol \NI.
The systems below assume the classical connectives.  
The third, stronger, system, {\bf P}, is the central system of this paper.
It has particularly appealing semantics.
The fourth system, {\bf CM}, is stronger than {\bf CL} but incomparable
with {\bf P}. It provides an example of a monotonic system that is
weaker than classical logic.
The last one of those systems, {\bf M}, is stronger than all previous systems
and equivalent to classical propositional logic. 

\section{The language, comparison with other systems} 
\label{sec:lan}
\subsection{Our language}
\label{subsec:lan}
The first step is to define a language in which to express the basic 
propositions. We shall assume that a set $L$ of well formed formulas 
(thereafter formulas) is given. It is very
important, from section~\ref{sec:pref} on, to assume that $L$ is closed under
the classical propositional connectives. They will be denoted by 
$\neg , \vee , \wedge , \rightarrow$ and $\leftrightarrow$.
Negation and disjunction will be considered as the basic connectives 
and the other ones as 
defined connectives. The connective $\rightarrow$ therefore denotes material
implication.
Small greek letters will be used to denote formulas.
Since no rule relating to the quantifiers will be discussed in this
paper, the reader may as well think of $L$ as the set of all propositional
formulas on a given set of propositional variables. 

With the language $L$, we assume semantics given by a set 
${\cal U}$, the elements of which will be referred to as worlds,
and a binary relation of {\em satisfaction} between worlds and formulas.
The set ${\cal U}$ is the universe of reference, it is the set of all 
worlds that we shall consider possible. If $L$ is the set of all
propositional formulas on a given set of propositional variables,
\cU\ is a subset of the set of all assignments of truth values to
the propositional variables.
We reserve to ourselves the right to consider universes
of reference that are strict subsets of the set of all models of $L$. 
In this way we shall be able to model {\em strict} constraints, such
as {\em penguins are birds}, in a simple and natural way, by restricting
\cU\ to the set of all worlds that satisfy the material implication
\mbox{$penguin \ra bird$}.
Typical universes of reference are given by the set of all propositional
worlds that satisfy a given
set of formulas. 

We shall assume that the satisfaction relation behaves as expected as far as
propositional connectives are concerned.
If $u\in {\cal U}$ and $\alpha, \beta \in L$
we write $u \models \alpha$ if $u$ satisfies $\alpha$ and assume:\\
1) $u \models \neg \alpha$ iff $u \not \models \alpha$. \\
2) $u \models \alpha \vee \beta$ iff $u \models \alpha$ or $u \models \beta$.

The notions of satisfaction of a set of formulas, validity of a 
formula and 
satisfiability of a set of formulas are defined as usual. 
We shall write $\models \alpha$ if 
$\alpha$ is valid, i.e. iff
$\forall u \in {\cal U}$, $u \models \alpha$, and write
\mbox{$\alpha \models \beta$} for
\mbox{$\models \alpha \ra \beta$}.

We shall also make the following 
{\bf assumption of compactness}\footnote{The compactness assumption is needed 
only to treat consequence relations defined as the set of all assertions
entailed by infinite sets of conditional 
assertions.}: a set of formulas is satisfiable if all of its
finite subsets are.

Classical theorems of compactness show that if we take $L$ to be a 
propositional calculus or a first order predicate calculus and \cU\ to be 
the set of all models that satisfy a given set of formulas, 
then the assumption
of compactness described above is satisfied. 
Notice that the set of valid formulas, in our sense, is not, in general, 
closed under substitutions. 

All that is done in the sequel depends on the choice of $L$ and ${\cal U}$,
though we shall often forget this dependence. For this work, the basic 
language $L$ may well be fixed, but we shall sometimes have to consider 
different universes of reference.
As noticed above, if $\Gamma$ is a set of formulas then
the subset of ${\cal U}$ that contains only the worlds that satisfy $\Gamma$
(this set of worlds will be denoted by ${\cal U}_\Gamma$) 
is a suitable universe. 

If $\alpha$ and $\beta$ are formulas, then the pair
\mbox{$\alpha \NI \beta$} 
(read {\em if $\alpha$, normally $\beta$}, or {\em \gb\ is a 
plausible consequence of \ga})
is called a conditional assertion (assertion in short). 
The formula \ga\ is the antecedent of the assertion, \gb\ is its
consequent.
The meaning we attach to such an assertion, and against which the reader 
should check the logical systems
to be presented in the upcoming sections, is the following: if 
$\alpha$  is true,
I am willing to (defeasibly)
jump to the conclusion that $\beta$ is true.
Our choice, then, is to look at {\em normally} as some binary notion.
It is clear that efforts to understand {\em normally} as some unary 
notion, e.g. translating {\em if \ga, normally \gb} as
\mbox{${\cal N} ( \alpha \ra \beta )$} or as \mbox{$\alpha \ra {\cal N}\beta$}
for some unary modal operator cannot be expressive enough.
{\em Consequence relations} are sets of conditional assertions. Not all such
sets, though, seem to be worthy of that name and our use of the term for
any such set is running against a fairly well-established terminology.
The term {\em conditional assertion} is taken from \cite{Scott:71} 
(p. 417).

We hope that, by considering nonmonotonic consequence as a meta-notion, but
allowing basic propositions on a rich language, we strike at the right 
language. It allows a new approach of questions about computational 
complexity (see \cite{Lewis:74} for some general 
decidability results), but this is left for future work.

\subsection{Pragmatics}
We shall now briefly sketch why we think that 
the study of nonmonotonic consequence relations will be a benefit to
the field of automated nonmonotonic reasoning.
The queries one wants to ask an automated knowledge base are formulas
(of $L$) and query \gb\ should be interpreted as:
{\em is \gb\ expected to be true}?
To answer such a query the knowledge base will apply some inference 
procedure to the information it has. We shall now propose a description
of the different types of information a knowledge base has.

The first type of information (first in the sense it is the more stable,
changes less rapidly) is coded in the universe of reference \cU,
that describes both hard constraints (e.g. dogs are mammals) and points
of definition (e.g. youngster is equivalent to not adult).
Equivalently, such information will be given by a set of formulas 
defining \cU\ to be the set of all worlds that satisfy all the formulas 
of this set.

The second type of information consists of a set of conditional assertions
describing the soft constraints (e.g. birds normally fly).
This set describes what we know about the way the world generally behaves.
This set of conditional assertions will be called the knowledge base,
and denoted by \bK.

The third type of information describes our information about the specific 
situation at hand (e.g. it is a bird). 
This information will be represented by a formula.

Our decision to consider the first type of information as
a separate type is not the only possible way to go.
One could, equivalently, treat a formula \ga\ of the 
first type as the conditional
assertion \mbox{$\neg$ \ga \NI {\bf false}}.
One could also have decided to introduce all information of the third type
as information of the first type.

Our inference procedure will work in the following way, to answer query
\gb.
In the context of the universe of reference \cU, 
it will try to deduce (in a way that is to be discovered yet)
the conditional assertion \ab\ from the knowledge base \bK.
This is a particularly elegant way of looking at the inference process:
the inference process deduces conditional assertions 
from sets of conditional assertions.
Clearly any system of nonmonotonic reasoning may be considered in this way.
So, we may look at circumscription, default logic and other systems as
mechanisms to deduce conditional assertions from sets of conditional 
assertions.
We shall now briefly investigate the expressive power of some of 
those systems in this light.

\subsection{Expressiveness of our language}
We shall now compare the expressive power of the language proposed
here, i.e. conditional assertions, to that of previous approaches.
Our purpose is to show that circumscription, autoepistemic logic and 
default logic all suffer from fundamental weaknesses, either in their
expressive capabilities or in their treatment of conditional information.
Let \ga, \gb\ and \gc\ be formulas.
We shall concentrate on the comparison of two different conditional
assertions. The conditional assertion \cA\ is 
\mbox{$\gamma \wedge \alpha$ \NI \gb}. The conditional assertion \cB\ is
\mbox{\gc \NI $\neg \alpha \vee \beta$}, i.e.
\mbox{\gc \NI $\alpha \ra \beta$}.
To simplify matters we shall just treat the special case 
when the formula \gc\ is a tautology. In this case
\cA\ is \ab\ and \cB\ is
\mbox{{\bf true} \NI $\neg \alpha \vee \beta$}.

The assertion \cA\ expresses that {\em if $\alpha$, normally $\beta$}.
Assertion \cB\ expresses that {\em Normally, if $\alpha$ is true then
$\beta$ is true}. Those assertions have very different meanings, at least
when $\alpha$ is normally false.
Assertion \cA\ says that in this exceptional case when $\alpha$ is true,
one also expects \gb\ to be true. Assertion \cB, on the other hand,
is automatically verified if $\alpha$ is normally false.
In any case it seems that it is perfectly possible that \cB\ does not
say anything about cases when \ga\ is true (if these are exceptional).
Take for example $\alpha$ to be {\em it is a penguin} and \gb\ to be
{\em it flies}. If we talk about birds, it seems perfectly 
reasonable to accept \cB\ which says that {\em normally, either it is not
a penguin or it flies}, since normally birds fly (and normally birds are
not penguins, but this remark is not necessary). Nevertheless, one
should be reluctant to accept \cA\ which says {\em penguins normally fly}.
It seems clear to us, then, that \cA\ and \cB\ have different meanings 
and that \cA\ does not follow from \cB. We agree with Y.~Shoham, and this
opinion will be supported in the sequel, to say that \cB\ should follow
from \cA, but we do not have to argue that case now.
In the main system to be presented in this paper, {\bf P},
the assertion \cA\ is strictly stronger than \cB. In the weaker
systems {\bf C} and {\bf CT}, \cA\ and \cB\ are incomparable.
In {\bf CM}, \cB\ is strictly stronger than \cA, and this is one of
the reasons we shall reject it as a system of nonmonotonic reasoning.
It is only in {\bf M}, which is equivalent to classical logic, 
that \cA\ and \cB\ are equivalent.

Let us consider now the expression of \cA\ and \cB,
first using circumscription.
For circumscription, \cA\ would be expressed as:
\mbox{$\alpha \wedge \neg {\em abnormal} \ra \beta$}.
In fact, since there would probably be a number of different abnormalities 
floating around, we probably should have written:
\mbox{$\alpha \wedge \neg {\em abnormal}_{543} \ra \beta$}, 
but this is not significant.
On the other hand \cB\ would be written as:
\mbox{$\neg {\em abnormal} \ra \left (\alpha \ra \beta \right )$}.
One immediately notices that the two formulations are logically equivalent.
We conclude that circumscription would need some additional mechanism
to distinguish between \cA\ and \cB.
In practise, the user of circumscription would give different priorities
(relative to the priorities of abnormalities of the other assertions
of the knowledge base), to the two abnormalities considered here;
but there is no standard procedure to determine priorities.

Let us now use autoepistemic logic.
The assertion \cA\ would be expressed as:
\mbox{$\alpha \wedge {\cal M} (\beta) \ra \beta$}. On the other hand
\cB\ would be expressed as:
\mbox{${\cal M} (\alpha \ra \beta) \ra \left ( \alpha \ra \beta \right )$}.
Since the modality \cM\ is interpreted as \mbox{$\neg$ \cK $\neg$}
for some epistemic modality \cK\ it satisfies:
\mbox{$\alpha \wedge {\cal M}(\beta) \ra {\cal M}(\alpha \ra \beta)$}.
We immediately notice that, for autoepistemic logic, \cB\ is strictly
stronger than \cA. This is not what we expect.

Let us try default logic now. The natural translation of \cA\ in
default logic would be the normal default:
\mbox{($\alpha , \beta ; \beta$)}, whose meaning is {\em if $\alpha$ has
been concluded to be true and $\beta$ is consistent with what has been
concluded so far, conclude that $\beta$ is true}.
The assertion \cB\ would be expressed as:
\mbox{(${\bf true} , \alpha \ra \beta ; \alpha \ra \beta$)}, which means
that in any situation in which $\alpha \ra \beta$ is consistent,
one should (or could) conclude this last formula to be true.
We immediately see that in all situations in which $\alpha$ has been
already concluded to be true, both defaults act exactly in the same way,
which seems very questionable.
In situations in which $\alpha$ has been concluded to be false, 
the first default is inapplicable, whereas the second default may
be applied but yields a trivial result (we do not get any new
information from applying it). 
Again, both defaults are equivalent, but, 
in this case, this seems to fit our intuition.
In situations in which neither $\alpha$ nor its negation have been
concluded, the first default cannot be applied. For the second default,
in certain situations it cannot be applied either, but in others it
may be applied and yields non trivial conclusions.
We conclude from this study that in some situations both defaults
are equivalent, in others the second is more powerful than the first one.
Again this is not what we expected. A particularly spectacular case
of this problem occurs
when $\beta$ is a logical contradiction.
The assertion 
\mbox{$\alpha \NI {\bf false}$} 
has a very clear
meaning: it says {\em if \ga, normally anything}. It expresses the
very strong statement that we are willing to disconsider completely the 
possibility of \ga\ being true. To see that this may express very useful
information, just think of \ga\ as {\em I am the Queen of England}.
Most people would probably be willing to have 
\mbox{$\alpha \NI {\bf false}$} 
in their personal knowledge base.
As remarked above, this corresponds to restricting \cU\ to those
worlds that do not satisfy \ga.
Now, the translation, as a normal default, of such an assertion,
which is:
\mbox{($\alpha , {\bf false} ; {\bf false}$)}, is never applicable since
{\bf false} is never consistent with anything. Therefore this default
gives no information at all. Somehow, all the strength of our assertion
has been lost in the translation. 

We hope to have convinced the reader that one should look for formalisms
in which the distinction between \cA\ and \cB\ is clear and understandable.

\section{Cumulative reasoning}

\subsection{Cumulative consequence relations}
\label{subsec:cum}
We shall, first, study the weakest of our logical systems. 
It embodies what we think, at this moment, 
in agreement with D.~Gabbay \cite{Gabbay:85}, 
are the rock-bottom properties without which 
a system should not be considered a logical system. 
This appreciation probably only reflects the fact that, 
so far, we do not know anything interesting about weaker systems.
The order of the exposition, roughly from weaker to stronger systems,
is aimed at minimizing repetitions:
rules that may be derived in weaker systems 
may also be derived in stronger ones.

We shall name this system {\bf C}, for {\em cumulative}. 
It is closely related to the cumulative
inference studied by D.~Makinson in \cite{Mak:89}, and
seems to correspond exactly, to what D.~Gabbay
proposed in \cite{Gabbay:85}. 
The system {\bf C} consists of a number of inference rules and an
axiom schema. 

\begin{definition}
\label{def:cumrel}
A consequence relation \NI\ is said to be {\em cumulative} iff it contains
all instances of the {\bf Reflexivity} axiom and is closed
under the inference rules of {\bf Left Logical Equivalence},
{\bf Right Weakening}, {\bf Cut} and {\bf Cautious Monotonicity} that will be
described below.
\end{definition}

We shall now describe and discuss the axioms and rules mentioned above 
and some derived rules. The purpose of the discussion is to weight
the meaning of the axioms and rules when the relation 
\mbox{$\ldots$ \NI\ $\ldots$} is interpreted
as {\em if $\ldots$ , normally $\ldots$}.

\begin{equation}
\label{ru:ref}
\alpha \NIm \alpha \hspace {2cm} {\rm ({\bf Reflexivity})}
\end{equation}

{\bf Reflexivity} seems to be satisfied universally by any kind of
reasoning based on some notion of consequence. 
Relations that do not satisfy it, probably express some notion of 
theory change.
It corresponds to the axiom ID of conditional logic.

The next two rules express the influence of the  
underlying logic, defined by the universe ${\cal U}$, 
on the notion of plausible consequence.
Their role is similar to that of the rules 
of consequence of \cite{Hoare:69}.

\begin {equation}
\label{ru:lle}
{{\models \alpha \leftrightarrow \beta \ \ , \ \  \alpha  \NIm \gamma} \over
    {\beta \NIm \gamma}} \hspace {2cm} 
{\rm ({\bf Left \ Logical \ Equivalence}) }
\end {equation}

{\bf Left Logical Equivalence} expresses the requirement that logically
equivalent formulas have exactly the same consequences
and corresponds to rule RCEA of conditional logic.
The consequences of a formula should depend on its meaning, 
not on its form. 
In the presence of the other rules of {\bf C}, 
it could have been weakened to: 
from \mbox{$\alpha \wedge \beta$ \NI $\gamma$} conclude
\mbox{$\beta \wedge \alpha$ \NI $\gamma$}.

The next rule, {\bf Right Weakening} expresses the fact that one 
must be ready to accept as plausible consequences all that is logically
implied by what one thinks are plausible consequences. In other words,
plausible consequences are closed under logical consequences.
It corresponds to the rule RCK of conditional logic.

\begin {equation}
\label{ru:rweak}
{{\models \alpha \ra \beta\ \ , \ \  \gamma \NIm \alpha } 
\over {\gamma \NIm \beta}} \hspace {2cm} {\rm ({\bf Right \ Weakening})}
\end {equation}

{\bf Right Weakening} obviously implies that one may replace logically
equivalent formulas by one another on the right of the \NI\ symbol.
{\bf Reflexivity} and {\bf Right Weakening} already imply that 
\mbox{$\alpha$ \NI $\beta$}\ if \mbox{$\alpha \models \beta$}.
All nonmonotonic systems proposed so far in the literature satisfy
{\bf Reflexivity}, {\bf Left Logical Equivalence} and {\bf Right
Weakening}.

Our next rule is named {\bf Cut} because of its similarity to Gentzen's 
{\em Schnitt}.

\begin {equation}
\label{ru:cut}
{{\alpha \wedge \beta \NI \gamma \ \ , \ \ \alpha \NI \beta} \over 
{\alpha \NI \gamma \;}} \hspace {2cm} {\rm ({\bf Cut})}
\end {equation}

It expresses the fact that one may, in his way towards
a plausible conclusion, first add an hypothesis to the facts he knows
to be true and prove the plausibility of his conclusion from this
enlarged set of facts and then deduce (plausibly)
this added hypothesis from the facts. This is a valid way of reasoning
in monotonic logic, and, as will be seen soon, its validity does not
imply monotonicity, therefore it seems to us quite reasonable to accept it.
Its meaning, it should be stressed, is that a plausible conclusion is
as secure as the assumptions it is based on. Therefore it may be added
(this is the origin of the term cumulative) into the assumptions.
There is no loss of confidence along the chain of derivations.
One may well be unwilling to accept such a principle and think that,
on the contrary, no conclusion of a derivation is ever as secure as
the assumptions.
Indeed, recently, D.~Gabbay~\cite{Gab:pers} suggested to replace
{\bf Cut} by a weaker rule. In this paper, we shall study only systems
that validate {\bf Cut}. Our conclusion is that there are many interesting
nonmonotonic systems that satisfy {\bf Cut}.
It should be mentioned that some probabilistic 
interpretations invalidate {\bf Cut} (Adams' validates it), 
e.g. interpreting a conditional assertion \ab\ as meaning 
that the corresponding conditional probability 
\mbox{$p(\beta \mid \alpha$)} is larger than some $q < 1$. 

It is easy to see that circumscription satisfies {\bf Cut}, at least
when all models that have to be considered are finite.
In~\cite{Mak:89}, D.~Makinson shows that default logic
satisfies {\bf Cut} too.
The following example should help convince the reader to endorse
{\bf Cut}. Suppose we tell you {\em we expect it will be raining
tonight} and {\em if it rains tonight, normally Fireball
should win the race tomorrow}. Wouldn't you conclude that we think 
that {\em normally, Fireball should win the race tomorrow}?  

Our last rule, named {\bf Cautious Monotonicity}, is taken from 
D.~Gabbay~\cite{Gabbay:85}.
It corresponds to axiom A3 of Burgess' system ${\cal S}$ in
\cite{Burgess:81}. 
The same property is named {\em triangulation} in 
\cite{PearlGeff:88}.

\begin {equation}
\label{ru:caut}
{{\alpha \NIm \beta \ \ , \ \ \alpha \NIm \gamma} 
\over {\alpha \wedge \beta
\NIm \gamma}}\hspace {2cm} {\rm ({\bf Cautious \  Monotonicity})}
\end {equation}

{\bf Cautious Monotonicity} expresses the fact that learning a new fact, 
the truth of which could have been plausibly concluded should
not invalidate previous conclusions.  It is a central property
of all the systems considered here.
The origin of the term {\em cautious monotonicity} will be explained in 
section~\ref{subsec:m}.
The probabilistic semantics that invalidates {\bf Cut} also invalidates
{\bf Cautious Monotonicity}. In~\cite{Mak:89}, D.~Makinson showed that
default logic, even when defaults are normal, does not always 
satisfy {\bf Cautious Monotonicity}. Circumscription, though, satisfies
it, at least when all models considered are finite.
What are our reasons to accept {\bf Cautious Monotonicity}?
On the general level, D.~Gabbay's argumentation seems convincing:
if \ga\ is reason enough to believe \gb\ and also to believe \gc,
then \ga\ and \gb\ should also be enough to make us believe \gc,
since \ga\ was enough anyway and, on this basis, \gb\ was expected.
From a pragmatic point of view {\bf Cautious Monotonicity} is very
important since we typically learn new facts and we would like to 
minimize the updating we have to make to our beliefs.
{\bf Cautious Monotonicity} and {\bf Cut} together tell us,
as will be made clear in lemma~\ref{le:cum}, that if the new facts 
learned were expected to be true, nothing changes in our beliefs.
This will help minimizing the updating.
From a semantic point of view, we want to argue the case for 
{\bf Cautious Monotonicity} on the following example. 
Suppose we tell you {\em we expect it will be raining
tonight} and {\em normally, Fireball
should win the race tomorrow}. Wouldn't you conclude that we think that
{\em even if it rains tonight, normally Fireball should win the race 
tomorrow}?  

\begin{lemma}
\label{le:cum}
The rules of {\bf Cut} and {\bf Cautious Monotonicity} may be expressed
together by the following principle: if \ab\ then the plausible consequences
of \ga\ and of \mbox{$\alpha \wedge \beta$} coincide.
\end{lemma}

Let us now consider some rules that may be derived in {\bf C}.

\subsection{Derived rules of {\bf C}}
\label{subsec:cumder}
The first rule corresponds to CSO of conditional logic and expresses
the fact that two propositions that are plausible consequences of each
other, have exactly the same plausible consequences.

\begin{equation}
\label{ru:equ}
{{\alpha \NIm \beta \ \ , \ \ \beta \NIm \alpha \ \ , \ \ 
\alpha \NIm \gamma} \over
{\beta \NIm \gamma}} \hspace{1.5cm}{\rm ({\bf Equivalence})}
\end{equation}

The second rule corresponds to CC of conditional logic and expresses
the fact that the conjunction of two plausible consequences 
is a plausible consequence.

\begin{equation}
\label{ru:and}
{{\alpha \NIm \beta \ \ , \ \ \alpha \NIm \gamma} \ \ \  \over
{\alpha \NIm \beta \wedge \gamma}} \hspace{3cm}{\rm ({\bf And})}
\end{equation}

The third rule amounts to {\em modus ponens} in the consequent.

\begin{equation}
\label{ru:mpcons}
{{\alpha \NIm \beta \ra \gamma \ , \ \alpha \NIm \beta} \over
{\alpha \NIm \gamma}} \hspace{2cm}{\rm ({\bf MPC})}
\end{equation}

The fourth rule is perhaps less expected and brought up here
to show that {\bf C} is not as weak as one could think.
It will be put to use in section~\ref{subsec:pref.char}.

\begin{equation}
\label{Or:Trans}
{{\ \ {{\alpha \vee \beta} \NI {\alpha}} \ \ , 
\ \ {\alpha \NI \gamma}} \ \ \ \over
{{\alpha \vee \beta} \NI \gamma}}
\end{equation}

\begin{lemma}
\label{le:dercum}
{\bf Equivalence}, {\bf And}, {\bf MPC} and (\ref{Or:Trans}) are derived 
rules of the system {\bf C}.
\end{lemma}
\proof
For {\bf Equivalence}, use first {\bf Cautious Monotonicity} to show that
\mbox{$\alpha \wedge \beta$ \NI $\gamma$}, then
{\bf Left Logical Equivalence} to get
\mbox{$\beta \wedge \alpha$ \NI $\gamma$} and then conclude by
{\bf Cut}.

For {\bf And}, first use {\bf Cautious Monotonicity} to show
\mbox{$\alpha \wedge \beta$ \NI $\gamma$}.
Then, since
\mbox{$\alpha \wedge \beta \wedge \gamma \models \beta \wedge \gamma$},
we have:
\mbox{$\alpha \wedge \beta \wedge \gamma$ \NI $\beta \wedge \gamma$}.
Using {\bf Cut} we conclude that
\mbox{$\alpha \wedge \beta$ \NI $\beta \wedge \gamma$}
and the desired conclusion is obtained by one more use of {\bf Cut}.

For {\bf MPC}, use {\bf And} and {\bf Right Weakening}.

For (\ref{Or:Trans}), remark that, since
\mbox{$\alpha \models \alpha \vee \beta$}, we have
\mbox{\ga \NI $\alpha \vee \beta$}. 
This, with the hypotheses,
enables us to apply {\bf Equivalence} and conclude.
\QED

\subsection{Monotonicity}
\label{subsec:m}
We shall now justify the term {\bf Cautious Monotonicity} and 
introduce four new rules. They cannot be derived in {\bf C}.
The first rule is {\bf Monotonicity}, or {\bf Left Strengthening}.

\begin{equation}
\label{ru:mon}
{{\models \alpha \ra \beta \ \ , \ \ \beta \NIm \gamma} \over
{\alpha \NIm \gamma}} \hspace {2cm} {\rm ({\bf Monotonicity})}
\end{equation}

It is clear that both {\bf Left Logical Equivalence} and
{\bf Cautious Monotonicity} are special cases of 
{\bf Monotonicity}.  This explains the name {\bf Cautious Monotonicity}.

The next rule corresponds to the {\em easy} half of the deduction theorem.

\begin {equation}
\label{ru:HD}
{{\alpha \NIm \beta \rightarrow \gamma} \ \ \ \over 
{\alpha \wedge \beta \NIm \gamma}} \hspace{2.2cm}{\rm ({\bf EHD})}
\end {equation} 

The next two rules have been considered by many.

\begin {equation}
\label{ru:trans}
{{\alpha \NIm \beta \ \ , \ \ \beta \NIm \gamma} 
\ \ \ \over
{\alpha \NIm \gamma}} \hspace{1.3cm}{\rm ({\bf Transitivity})}
\end {equation}

\begin{equation}
\label{ru:contra}
{{\alpha \NIm \beta} \ \ \ \ \over
{\neg \beta \NIm \neg \alpha}} \hspace{3cm}{\rm ({\bf Contraposition})}
\end{equation}

It is easy to find apparent counter-examples 
to each one of the last four rules
in the
folklore of nonmonotonic reasoning. 
The next lemmas will explain why. Let us notice that, nevertheless,
adding the first three of those rules to the system {\bf C} leaves us with a 
system, 
{\bf CM}, that is strictly weaker than classical monotonic logic, as
will be seen in section~\ref{sec:mon.cum}. 
The next lemmas will describe some of the relations between
the rules above.

\begin{lemma}
\label{le:dermon}
In the presence of the rules of {\bf C}, the rules of {\bf Monotonicity},
{\bf EHD}, and {\bf Transitivity} are all equivalent.
\end{lemma}
\proof
We shall not mention the uses of {\bf Reflexivity}, {\bf Left Logical 
Equivalence} and {\bf Right Weakening}.
{\bf Monotonicity} implies {\bf EHD}, using {\bf And}.
{\bf EHD} implies {\bf Monotonicity}.
{\bf Monotonicity} implies {\bf Transitivity}, using {\bf Cut}.
{\bf Transitivity} implies {\bf Monotonicity}.
\QED

\begin{lemma}
\label{le:dermon2}
In the presence of {\bf Left Logical Equivalence} and {\bf Right Weakening},
{\bf Contraposition} implies {\bf Monotonicity}.
\end{lemma}
\proof
Use {\bf Contraposition}, then {\bf Right Weakening} and {\bf Contraposition}
again.
\QED

The results of section~\ref{sec:mon.cum} show that {\bf Monotonicity} does
not imply {\bf Contraposition} even in the presence of the rules of {\bf C}.
Since {\bf Monotonicity} seems counter-intuitive in nonmonotonic systems,
the two lemmas above show we should not accept {\bf EHD}, 
{\bf Transitivity} or {\bf Contraposition}.

\subsection{Cumulative models}
We shall now develop a semantic account of cumulative reasoning,
i.e. reasoning using the rules of the system {\bf C}.
We shall define a family of models
(without any reference to the rules of {\bf C}) 
and show how each model defines a consequence relation. 
We shall then show that each model of the family
defines a cumulative consequence relation (this is a soundness result)
and that every cumulative consequence relation is defined by some model 
of the family (this is a completeness result, or a representation 
theorem).

Let us, first, describe the models informally. A model essentially consists
of a set of states (they represent possible states of affairs, including
perhaps the state of mind or knowledge of the reasoner)
and a binary relation on those states. The relation 
represents the preferences the reasoner may have between different states: 
he could for example
prefer the states in which he is  rich to the ones in which he is poor,
and prefer the states in which he knows he is rich to those in which he is
rich but does not know about it. 
More realistically, one could prefer states
in which Tweety is a bird and flies to those in which Tweety is a bird but
does not fly.
The reasoner, described by a model, accepts a conditional assertion
\ab\ iff all those states that are {\em most preferred} among all
states satisfying \ga, satisfy \gb.
The reader should notice we have not yet said what is a state
and what formulas are satisfied by a state. 

We shall not define further the notion of a state, 
but suppose that every state is, in a model, labeled with
a set of worlds (intuitively the set of all worlds the reasoner
thinks are possible in this state).
Modal logicians will identify our labels as S5 models.
Considering a binary relation on states labeled by sets of worlds,
instead of considering a binary relation on sets of worlds, 
gives us an additional degree of freedom in building models:
the same set of worlds may appear at many states (that are not
equivalent from the point of view of the binary relation).
This additional freedom is vital for the representation theorem to
hold, and was missing from Shoham's account~\cite{Shoham:87}.

Some technical definitions are needed first.

\begin{definition}
\label{def:asy}
Let $\prec$ be a binary relation on a set $U$. We shall say that
$\prec$ is {\em asymmetric} iff 
\mbox{$\forall s , t \in U$} such that 
\mbox{$s \prec t$}, we have
\mbox{$t \not \prec s$}.
\end{definition}

\begin{definition}
\label{def:min}
Let \mbox{$V \subseteq U$} and $\prec$ a binary relation on $U$.
We shall say that 
\mbox{$t \in V$} 
is {\em minimal} in $V$ iff
\mbox{$\forall s\in V$},
\mbox{$s \not \prec t$}.
We shall say that
\mbox{$t \in V$} 
is a {\em minimum} of $V$ iff \/
\mbox{$\forall s\in V$} such that \mbox{$s \neq t$},
\mbox{$t \prec s$}.
\end{definition}

The reader has noticed that, though the last definitions sound familiar 
in the case
the relation $\prec$ is a strict partial order, we intend to use them for 
arbitrary relations.

\begin{definition}
\label{def:smoo}
Let \mbox{$P \subseteq U$} and $\prec$ a binary relation on $U$.
We shall say that $P$ is  {\em smooth} iff
\mbox{$\forall t \in P$}, either 
\mbox{$\exists s$} minimal in $P$, such that \mbox{$s \prec t$} or
$t$ is itself minimal in $P$.
\end{definition}

We shall use the following lemma, the proof of which is obvious.

\begin{lemma}
\label{min:min}
Let $U$ be a set and $\prec$ an asymmetric binary relation on 
$U$. If\ $U$ has a minimum it is unique, it is a minimal element of $U$ and
$U$ is smooth.
\end{lemma}

\begin{definition}
\label{def:cum.mod}
A {\em cumulative} model is a triple 
\mbox{$\langle \: S , l , \prec \rangle$},
where $S$ is a set, the elements of which are called states, 
\mbox{$l:S \mapsto 2^{\,\cal U}$} is a function that labels every state 
with a non-empty set of worlds and $\prec$ 
is a binary 
relation on $S$, 
satisfying the {\bf smoothness condition} that will be defined below
in definition~\ref{def:smoocond}.
\end{definition}

The relation $\prec$ represents the reasoner's preference among states.
The fact that $s \prec t$ means that, in the agent's mind, 
$s$ is {\em preferred} to or more {\em natural} than $t$. 
As will be formally defined below,
the agent is willing to conclude
$\beta$ from $\alpha$, if all {\em most natural} states which satisfy $\alpha$
also satisfy $\beta$.

\begin{definition}
\label{def:EM}
Let \mbox{$\langle \: S , l , \prec \rangle$} be as above.
If \ga\ is a formula, 
we shall say that $s \in S$ satisfies \ga\ and write
\mbox{$s \EM \alpha$} iff for every world \mbox{$m \in l(s)$}, 
\mbox{$m \models \alpha$}.
The set:
\mbox{$\{ s \mid s \in S, \: s \EM \alpha \}$} 
of all states that satisfy \ga\ will be denoted by
\mbox{$\widehat{\alpha}$}.
\end{definition}

\begin{definition}[smoothness condition]
\label{def:smoocond}
A triple
\mbox{$\langle \: S , l , \prec \rangle$} is said to satisfy
the smoothness condition iff,
\mbox{$\forall \alpha \in L$}, the set 
\mbox{$\widehat\alpha$} is smooth.
\end{definition}

The smoothness condition is necessary to ensure the validity of
{\bf Cautious Monotonicity}. It is akin to the {\em limit assumption}
of Stalnaker~\cite{Stal:68} and Lewis~\cite{Lewis:73}, but it is defined
in a more general context. Smoothness is the property called,
contrary to mathematical usage,
{\em well-foundedness} in~\cite{Eth:85} and in~\cite{Lif:86}.

We shall now describe how a cumulative model defines a consequence relation.

\begin{definition}
\label{def:cumcons}
Suppose a cumulative model \mbox{$W = \langle S , l, \prec \rangle$}
is given. The consequence relation 
defined by $W$ will be denoted by \NIW\ and is defined by:
\mbox{\ga \NIW \gb} iff for any $s$ minimal in $\widehat{\alpha}$,  
\mbox{$s \EM \beta$}. 
\end{definition}

\begin{definition}
\label{def:strongcum}
A triple 
\mbox{$\langle \: S , l , \prec \rangle$} 
is said to
be a {\em strong} cumulative model iff
\begin{enumerate}
\item the relation $\prec$ is asymmetric
\item for each formula \ga, the set $\widehat\alpha$ has a minimum.
\end{enumerate}
\end{definition}

It is clear that strong cumulative models are cumulative models, 
i.e. satisfy the smoothness condition.
The definition of cumulative models and the consequence relations they 
define seems quite natural, i.e. a preference relation on epistemic 
states, but one should not forget that the preference relation $\prec$
is not required to be a partial order and that 
in triples (even when the set of states is finite) in which 
the relation $\prec$ is
not a partial order, the smoothness condition is, in general, not
an easy thing to check.

\subsection{Characterization of cumulative relations}
\label{subsec:char.cumul}

In this section we shall characterize the relation 
between cumulative consequence
relations and cumulative models.
The first lemma is obvious.

\begin{lemma}
\label{remark1}
Let \mbox{$W = \langle \: S , l , \prec \rangle$} be a cumulative model.
For \mbox{$\alpha , \beta \in L$},
\mbox{${\widehat{\alpha \wedge \beta}} =$} 
\mbox{${\widehat{\alpha} \cap \widehat{\beta}}$}.
\end{lemma}

\begin{lemma}[Soundness]
\label{Cu:Sou}
For any cumulative model $W$, the consequence relation \NIW\ it
defines is a cumulative relation, i.e. all the rules
of the system {\bf C} are satisfied by the relations defined by 
cumulative models.
\end{lemma}
\proof
The proof is easy and we shall only treat {\bf Cut} and 
{\bf Cautious Monotonicity}.
The smoothness condition is needed only for dealing with
{\bf Cautious Monotonicity}.

For {\bf Cut}, suppose all minimal elements of $\widehat\alpha$ satisfy
\gb\ and all minimal elements of \mbox{$\widehat{\alpha \wedge \beta}$}
satisfy \gc. Any minimal element of $\widehat\alpha$ satisfies \gb\ and
therefore satisfies $\alpha \wedge \beta$. Since it is minimal in 
$\widehat\alpha$ and 
\mbox{$\widehat{\alpha \wedge \beta} \subseteq \widehat{\alpha}$}, 
it is also minimal in \mbox{$\widehat{\alpha \wedge \beta}$}.

For {\bf Cautious Monotonicity}, the smoothness condition is needed.
Suppose that \mbox{$\alpha \NIW \beta$} and \mbox{$\alpha \NIW \gamma$}. 
We have to prove that
\mbox{$\alpha \wedge \beta \NIW \gamma$}, i.e., that, for any $s$ minimal in 
\mbox{$\widehat{\alpha \wedge \beta}$}, \mbox{$s \EM \gamma$}.
Such an $s$ is in $\widehat{\alpha}$. We shall prove that it is minimal in 
$\widehat{\alpha}$. By the smoothness condition, if it were not minimal,
there would be an $s'$ minimal in $\widehat{\alpha}$ such that
\mbox{$s' \prec s$}. But \mbox{$\alpha \NIW \beta$} therefore 
\mbox{$s' \EM \beta$} and then 
\mbox{$s' \in {\widehat\alpha \cap \widehat\beta}$}.
By lemma~\ref{remark1} we conclude that $s'$ is in 
\mbox{$\widehat{\alpha \wedge \beta}$}, 
in contradiction with the minimality of $s$ in this set.
Therefore $s$ is minimal in \mbox{$\widehat{\alpha}$} and, since 
\mbox{$\alpha \NIW \gamma$},
one concludes: \mbox{$s \EM \gamma$}.
\QED

We now intend to show that, given any cumulative relation 
\NI, one may build a cumulative model $W$ that defines a 
consequence relation
\mbox{$\NIW$} that is exactly \mbox{$\NI$}. 
Suppose, therefore, that \NI\ satisfies the rules
of {\bf C}. 
All definitions will be relative to this relation. 

\begin{definition}
\label{def:normworld}
The world \mbox{$m \in {\cal U}$} is said to be a {\em normal} 
world for $\alpha$ iff 
\mbox{$\forall \beta \in L$} such that \mbox{$\alpha$ \NI $\beta$}, 
\mbox{$m \models \beta$}.
\end{definition}

So, a world is normal for a formula if it satisfies all of its
plausible consequences.
Obviously, if the consequence relation we start from satisfies 
{\bf Reflexivity}, a normal world for $\alpha$ satisfies $\alpha$.

\begin{lemma}
\label{norm:mod}
Suppose a consequence relation \NI\ satisfies {\bf Reflexivity}, 
{\bf Right Weakening} and {\bf And}, 
and let \mbox{$\alpha , \beta \in L$}.
All normal worlds for $\alpha$  satisfy $\beta$ iff 
\mbox{$\alpha$ \NI $\beta$}.
\end{lemma}
\proof
The {\em if} part follows from definition \ref{def:normworld}.
Let us show the {\em only if} part.
Suppose \mbox{$\alpha$ \notNI $\beta$}, 
we shall build a normal world for 
$\alpha$ that does not satisfy $\beta$.
Let \mbox{$ \Gamma_0 \eqdef$}
\mbox{$ \{ \neg \beta \} \cup \{\delta \mid \alpha \NIm \delta \}$}.
It is enough to show that $\Gamma_0$ is satisfiable.
Suppose not, then, by the compactness assumption, there
exists a finite subset of $\Gamma_0$ that is not satisfiable and therefore
a finite set \mbox{$D \subseteq \{\delta \mid \alpha \NI \delta\}$} 
such that
\mbox{$\models \bigwedge_{\delta \in D} \delta \rightarrow \beta $}.
Now, 
\mbox{$\models \alpha \rightarrow \left ( \bigwedge_{\delta \in D} \delta 
\rightarrow \beta \right )$}
and, by {\bf Reflexivity} and {\bf Right Weakening}
\mbox{$\alpha$ \NI $\left ( \bigwedge_{\delta \in D} \delta \rightarrow 
\beta \right )$}.
But, using {\bf And} one gets 
\mbox{$\alpha \NI \bigwedge_{ \delta \in D} \delta$}.
Then, using {\bf MPC} (the proof of lemma~\ref{le:dercum} shows that
only {\bf And} and {\bf Right Weakening} are needed to derive {\bf MPC}),
one concludes \mbox{$\alpha \NI \beta$}, 
a contradiction. 
\QED

\begin{definition}
\label{tildeq}
We shall say that $\alpha$ is {\em equivalent} to $\beta$ and write 
\mbox{$\alpha \sim \beta$} iff \mbox{$\alpha$ \NI $\beta$} and 
\mbox{$\beta$ \NI $\alpha$}.
\end{definition}

\begin{lemma}
\label{le:eqcu}
\mbox{$\alpha \sim \beta$} iff \/ $ \forall \gamma \ \  \alpha \NIm \gamma 
\Leftrightarrow \beta \NIm \gamma$. The relation $\sim$ is therefore an
equivalence relation.
\end{lemma}
\proof
The {\em if} part follows from {\bf Reflexivity} and
the {\em only if} part from {\bf Equivalence}.
\QED

The equivalence class of a formula $\alpha$, under $\sim$, 
will be denoted by $\bar{\alpha}$.

\begin{definition}
\label{def:leq}
\mbox{$\bar{\alpha} \leq \bar{\beta}$} iff 
\mbox{$\exists \alpha' \in \bar{\alpha}$}
such that 
\mbox{$\beta$ \NI $\alpha'$}.
\end{definition}

It is clear that the definition of $\leq$ makes sense, i.e. does not depend
on the choice of the representatives $\alpha$ and $\beta$.
The relation $\leq$ is reflexive but is not in general transitive.

\begin{lemma}
\label{asymm}
The relation $\leq$ is antisymmetric.
\end{lemma}
\proof
Suppose \mbox{$\bar{\alpha} \leq \bar{\beta}$} and 
\mbox{$\bar{\beta} \leq \bar{\alpha}$}.
There are formulas \mbox{$\alpha' , \alpha'' \in \bar{\alpha}$} and 
\mbox{$\beta' , \beta'' \in \bar{\beta}$} such that:
\mbox{$\beta' \NIm \alpha''$} and \mbox{$\alpha' \NIm \beta''$}.
By lemma~\ref{le:eqcu},
\mbox{$\beta'' \NIm \alpha''$} and 
\mbox{$\alpha'' \NIm \beta''$}. 
Therefore \mbox{$\alpha'' \sim \beta''$}, and 
\mbox{$\bar{\alpha} = \bar{\beta}$}.
\QED

The cumulative model $W$ will be defined the following way:
\mbox{$W \eqdef \langle \: S , l , \prec \rangle$}, where
\mbox{$S \eqdef L / \!\! \sim$} 
is the set of all equivalence classes of formulas
under the relation $\sim$,
\mbox{$l ( \bar{\alpha} ) \eqdef$}
\mbox{$\{ m \mid m $\mbox{ is a normal world for }$ \alpha \}$}
and 
\mbox{$\bar{\alpha} \prec \bar{\beta}$} iff 
\mbox{$\bar{\alpha} \leq \bar{\beta}$}
and \mbox{$\bar{\alpha} \neq \bar{\beta}$}
(the relation $\leq$ has been defined in definition~\ref{def:leq}).
One easily checks the definition of $l$ does not depend on the choice of
the representative $\alpha$ and that $\prec$ is asymmetric.

\begin{lemma}
\label{le:minimum}
For any \mbox{$\alpha \in L$} the state $\bar{\alpha}$ is a minimum
of $\widehat{\alpha}$. 
\end{lemma}
\proof
Indeed suppose \mbox{$s \neq \bar{\alpha}$} and 
\mbox{$s \in \widehat{\alpha}$}. 
This last assumption implies, by the definition
of $\widehat{\alpha}$, that every world
of \mbox{$l ( s )$} satisfies $\alpha$.
Let \mbox{$s = \bar{\beta}$}. By the definition of $l$,
every normal world for $\beta$ satisfies $\alpha$. By lemma~\ref{norm:mod},
\mbox{$\beta \NI \alpha$}, and therefore
\mbox{$\bar{\alpha} \leq s$}.
Since \mbox{$s \neq \bar{\alpha}$} we conclude 
\mbox{$\bar{\alpha} \prec s$}.
\QED

It follows from lemma~\ref{le:minimum} and the fact that $\prec$
is asymmetric that the model $W$ defined above is a strong cumulative model.
We may now prove what we wanted to achieve. 
\begin{lemma}
\label{un}
\mbox{$\alpha \NI \beta$} iff \mbox{$\alpha \NIW \beta$}.
\end{lemma}
\proof
Lemmas~\ref{le:minimum} and \ref{min:min} imply that 
the only minimal state of 
$\widehat{\alpha}$ is $\bar{\alpha}$, therefore 
\mbox{$\alpha \NIW \beta$}
iff all normal worlds for $\alpha$ satisfy $\beta$ and lemma~\ref{norm:mod}
implies the conclusion.
\QED

\begin{theorem}[Representation theorem for cumulative relations]
\label{th:repcum}
A consequence relation is a cumulative consequence relation iff it is defined
by some cumulative model.
\end{theorem}
\proof
The {\em if} part is lemma~\ref{Cu:Sou}.
The {\em only if} part follows from the construction of $W$ 
and lemma~\ref{le:minimum} (that shows $W$ is a cumulative model)
and lemma~\ref{un}.
\QED

One may remark that the representation result proved is a bit stronger than
what is claimed in theorem~\ref{th:repcum}: 
any cumulative consequence relation is the consequence relation
defined by a strong cumulative model.
It is now easy to study the notion of entailment
yielded by cumulative models.

\begin{corollary}
\label{log:imp3}
Let {\bf K} be a set of conditional assertions, and 
\mbox{$\alpha , \beta \in L$},
the following conditions are equivalent. In case they hold we shall say
that {\bf K} cumulatively entails \ab.
\begin{enumerate}
\item for all cumulative models $V$ such that \mbox{$\NIV$} contains {\bf K}, 
$\alpha \NIV \beta$
\item \mbox{$\alpha \NI \beta$}  has a proof from {\bf K} 
in the system {\bf C}.
\end{enumerate}
\end{corollary}
\proof
From lemma~\ref{Cu:Sou} one sees that 2) implies 1).
For the other direction, suppose 2) is not true. The smallest 
consequence relation closed under the rules of {\bf C} that contains
{\bf K} is a cumulative consequence relation that does not contain
\ab. By theorem~\ref{th:repcum}, there is a cumulative model that
defines it. This model shows property 1) does not hold.
\QED

\begin{corollary}
\label{co:cum2}
Let {\bf K} be a set of conditional assertions. 
There is a cumulative model that satisfies exactly
those assertions that are cumulatively entailed by
\bK.
\end{corollary}

The following compactness result follows.
\begin{corollary}[compactness]
\label{comp:cum}
{\bf K} entails
\ab\ iff a finite subset of {\bf K} does.
\end{corollary}
\proof
Proofs are always finite and therefore use only a finite number
of assumptions from {\bf K}.
\QED

To conclude our study of cumulative reasoning, let us say that the
system {\bf C} provides an interesting general setting in which
to study nonmonotonic reasoning. Weaker systems are probably very
different from systems that are at least as strong as {\bf C}.
The system {\bf C} is probably too weak to be the backbone of realistic
inference systems and cumulative models are quite cumbersome to
manipulate. The next section will propose nicer models and an
additional rule of inference.

\section{Cumulative reasoning with Loop}
\label{sec:cum.loop.reas}
\subsection{Cumulative ordered models}
\label{subsec:cum.ord}
The original motivation for the study of the system {\bf CL}, that will
be proposed
in this section, stems from semantic considerations. 
Later on, a number of results, including the result that will be 
described in section~\ref{subsec:horn},
which says that, if one restricts oneself to Horn assertions, then
the system {\bf CL} is as strong as {\bf P}, seemed to point out 
that {\bf CL} is worth studying.

Looking back on the cumulative models of definition~\ref{def:cum.mod},
one may wonder why we did not require the binary relation $\prec$ to be
a strict partial order. We could have required it to be asymmetric without
jeopardizing the representation theorem,
but the construction of
section~\ref{subsec:char.cumul} builds a model in which $\prec$ is not
always transitive. Nevertheless, preferences could probably be assumed to be 
transitive and, most important, transitivity of $\prec$ eases
enormously the task of checking the smoothness condition: if $\prec$ is
a partial order (strict), then all finite models (models in which the set
of states is finite) satisfy the smoothness condition, and even all 
well-founded models 
(in which there is no infinite descending $\prec$-chain) do.
Could we have required $\prec$ to be a partial order? 
In other terms, are there rules that are
not valid for cumulative models in general but are valid for all cumulative
models the preference relation of which is a strict partial order?
We shall now give a positive answer to this last question and exactly 
characterize this sub-family of cumulative models.

\begin{definition}
A cumulative {\em ordered} model is a cumulative model in which the relation
$\prec$ is a strict partial order.
\end{definition}

\subsection{The system {\bf CL}}
The following rule, named {\bf Loop} after its form, will be shown
to be the exact counterpart of transitivity of the preference relation
in the models.
It says that, if propositions may be arranged in a loop, in a way
each one is a plausible consequence of the previous one, then each
one of them is a plausible consequence of any one of them, 
i.e. they are all equivalent in the sense of {\bf Equivalence}.

\begin{definition}
\label{def:CL}
The system {\bf CL} consists of all the rules of {\bf C} and the following:
\begin{equation}
\label{loop}
{{\alpha_0 \NIm \alpha_1 \ \ , \ \ \alpha_1 \NIm \alpha_2 
\ \ , \ldots ,\ \ \alpha_{k-1} \NIm \alpha_k
\ \ , \ \ \alpha_k \NIm \alpha_0} \over 
{\alpha_0 \NIm \alpha_k}}
\hspace{1.5cm}{\bf (Loop)}
\end{equation}
A consequence relation that satisfies all rules of {\bf CL} is
said to be {\em loop-cumulative}.
\end{definition}

\begin{lemma}
\label{le:derloop}
The following is a derived rule of {\bf CL}, for any 
\mbox{$i , j = 0 , \ldots k$}.
\begin{equation}
\label{ru:derloop}
{{\alpha_0 \NIm \alpha_1 \ \ , \ \ \alpha_1 \NIm \alpha_2 
\ \ , \ldots ,\ \ \alpha_{k-1} \NIm \alpha_k
\ \ , \ \ \alpha_k \NIm \alpha_0} \over 
{\alpha_i \NIm \alpha_j}}
\end{equation}
\end{lemma}
\proof
It is clear that, because of the invariance of the assumptions under
cyclic permutation, the conclusion of {\bf Loop}, could as well have been 
\mbox{$\alpha_{i+1}$ \NI $\alpha_{i}$}, for any $i = 0 , \ldots , k$
(addition is understood modulo $k+1$).
From {\bf Equivalence} one can then conclude 
\mbox{$\alpha_i$ \NI $\alpha_{j}$}, for any $i , j = 0 , \ldots , k$.
\QED

It seems the rule {\bf Loop} has never been considered in the literature.
We feel it is an acceptable principle of nonmonotonic reasoning.
It is particularly interesting that {\bf Loop} does not mention any
of the propositional connectives.

\begin{lemma}
\label{val:loop}
{\bf Loop} is valid in all cumulative ordered models.
\end{lemma}
\proof
Let \mbox{$W = \langle S , l , \prec \rangle$} 
be a cumulative ordered model
such that \mbox{$\alpha_i$ \NIW $\alpha_{i + 1}$} for 
\mbox{$i = 0 , \ldots , k$} 
(addition is understood modulo $k+1$) and 
let \mbox{$s_0 \in S$} be a minimal state in $\widehat{\alpha_0}$.
We shall show that \mbox{$s_0 \EM \alpha_k$}.
Since \mbox{$\alpha_0$ \NIW $\alpha_1$}, the state 
$s_0$ must be in $\widehat{\alpha_1}$.
By the smoothness condition, if $s_0$ is not minimal in $\widehat{\alpha_1}$
then there is a state $s_1$ minimal in 
$\widehat{\alpha_1}$ such that \mbox{$s_1 \prec s_0$}.
Similarly, for every \mbox{$i = 0 , \dots , k$} there is a state $s_i$ 
minimal in $\widehat{\alpha_i}$ such that \mbox{$s_i = s_{i-1}$} or 
\mbox{$s_i \prec s_{i - 1}$}.
Since $\prec$ is transitive, \mbox{$s_k = s_0$} or \mbox{$s_k \prec s_0$}.
But $s_k$ is minimal in $\widehat{\alpha_k}$ and 
\mbox{$\alpha_k$ \NIW $\alpha_0$},
we conclude that $s_k \in \widehat{\alpha_0}$.
But $s_0$ is minimal in $\widehat{\alpha_0}$,
we conclude that \mbox{$s_k = s_0$} and \mbox{$s_0 \EM \alpha_k$}.
\QED

\begin{lemma}
\label{notCvalid}
The rule {\bf Loop} is not valid in cumulative models.
\end{lemma}
\proof
Let $L$ be the propositional calculus on the propositional variables
$p_0 , p_1 , p_2$ and ${\cal U}$ be the set of all 
propositional models on those
variables.
We shall build a cumulative model \mbox{$V = \langle S , l , \prec \rangle$} 
such that \mbox{$p_i$ \NIV $p_{i + 1}$} 
for all $i = 0 , \dots , 2$ (addition is modulo 3) but 
\mbox{$p_0 \notNIV p_2$}.
The set $S$ has four states: $s_i$, for $i = -1 , \ldots , 2$.
For every $i = 0 , \dots , 2$ we have \mbox{$s_{-1} \prec s_i$} and 
\mbox{$s_{i + 1} \prec s_i$}.
Notice that $\prec$ is not transitive.
Let us now describe $l$. For $i = 0 , \ldots , 2$, $l ( s_i )$ is the set
of all worlds satisfying $p_i$ and $p_{i + 1}$, and $l ( s_{-1} )$ is the set
of all worlds satisfying at least two out of the three variables.
First we want to show that $V$ satisfies the smoothness condition.
Clearly all subsets of $S$ that contain $s_{-1}$ are smooth since $s_{-1}$
is a minimum in $S$. A set that contains at most two elements is
always smooth. We conclude that the only subset of $S$ that is not smooth
is \mbox{$A \eqdef \{ s_0 , s_1 , s_2 \}$}. 
We must show that there is no formula
$\alpha$ such that \mbox{$A = \widehat{\alpha}$}. Let $\alpha$ be any formula 
and let $i = 0 , \ldots , 2$.
If \mbox{$s_i \in \widehat{\alpha}$} all worlds of 
\mbox{$l ( s_i )$} must satisfy $\alpha$
and by definition of $l$, \mbox{$p_i \wedge p_{i + 1} \models \alpha$}.
We conclude that if \mbox{$A \subseteq \widehat{\alpha}$} then 
any world that satisfies at least two of the variables satisfies $\alpha$.
We conclude that $\widehat{\alpha}$ must therefore also include $s_{-1}$.

To see that \mbox{$p_i$ \NIV $p_{i + 1}$}, notice that 
\mbox{$\widehat{p_i} = \{ s_{i - 1} , s_i \}$} 
and that, since \mbox{$s_i \prec s_{i-1}$},
the only minimal state in $\widehat{p_i}$ is $s_i$ that satisfies $p_{i + 1}$.
The only thing left to check is that \mbox{$p_0$ \notNIV $p_2$}.
But we just noticed that the only minimal state of
$\widehat{p_0}$ is $s_0$ and clearly \mbox{$s_0 \notEM p_2$}.
\QED

\subsection{Characterization of loop-cumulative consequence relations}
We now want to show that, given any loop-cumulative relation \NI\ 
one may build a cumulative ordered model $V$ such that
\NIV\ is equal to \NI.
Suppose \NI\ is such a relation and 
\mbox{$W = \langle S , l , \prec \rangle$}  is 
the cumulative model built out
of \NI\ in section~\ref{subsec:char.cumul}.
Let $\prec^+$ be the transitive closure of $\prec$.
First we shall show that, since \NI\ satisfies {\bf Loop}, the relation
$\prec^+$ is a strict partial order.
\begin{lemma}
\label{strictpo}
The relation $\prec^+$ is irreflexive and therefore a strict partial order.
\end{lemma}
\proof
Suppose \mbox{${\bar{\alpha}}_0 \prec^+ {\bar{\alpha}}_0$}.
Since $\prec$ is asymmetric, it is irreflexive and t
here must be some \mbox{$n > 0$} such that for 
\mbox{$i = 0 , \ldots , n$},
\mbox{${\bar{\alpha}}_i \prec {\bar{\alpha}}_{i+1}$}
(addition is modulo $n$).
From the definitions of $\prec$ and $\leq$, we see that, for 
\mbox{$i = 0 , \ldots , n$},
there are formulas $\alpha'_i$ such that
\mbox{$\alpha_i \sim \alpha'_i$} and \mbox{$\alpha_{i+1}$ \NI $\alpha'_i$}.
From lemma~\ref{le:eqcu}, we conclude that
\mbox{$\alpha'_{i+1}$ \NI $\alpha'_i$} for
\mbox{$i = 0 , \ldots , n$}.
By {\bf Loop} we see that \mbox{$\alpha'_i$ \NI $\alpha'_{i+1}$}
and therefore \mbox{$\alpha'_{i+1} \sim \alpha'_i$} and 
\mbox{$\alpha_i \sim \alpha_{i+1}$}.
But this contradicts the asymmetry of $\prec$. We have shown that
$\prec$ is irreflexive.
Since it is transitive by construction it is a strict partial order.
\QED

Let us now define \mbox{$V \eqdef \langle S , l , \prec^+ \rangle$}
where $S , l $ and $ \prec $ are as in the definition of $W$.
\begin{lemma}
\label{V:min}
In $V$, for any $\alpha$, the state $\bar{\alpha}$ is a minimum of 
$\widehat{\alpha}$. Therefore $V$ is a strong cumulative ordered model.
\end{lemma}
\proof
Lemma~\ref{le:minimum} says $\bar{\alpha}$ is a minimum of $\widehat{\alpha}$
with respect to $\prec$. It is therefore a minimum with respect to 
any weaker relation
and in particular $\prec^+$.
Lemma~\ref{strictpo} implies that $\prec^+$ is asymmetric and, by
lemma~\ref{min:min}, $V$ satisfies the smoothness condition.
\QED

\begin{lemma}
\label{iff}
\ab\ iff \mbox{$\alpha$ \NIV $\beta$}.
\end{lemma}
\proof
Lemma~\ref{V:min} implies that the only minimal state of 
$\widehat{\alpha}$ is $\bar{\alpha}$, therefore \mbox{$\alpha$ \NIV $\beta$}
iff all normal worlds for \ga\ satisfy \gb, and lemma~\ref{norm:mod}
implies the conclusion.
\QED

We may now summarize.
\begin{theorem}[Representation theorem for loop-cumulative relations]
\label{rep:PC}
A consequence relation is a loop-cumulative relation
iff it is defined by some cumulative ordered model.
\end{theorem}

As in the cumulative case one may study the notion of 
entailment yielded by cumulative ordered models and obtain
results that parallel corollaries~\ref{log:imp3}, \ref{co:cum2} 
and \ref{comp:cum}.

\section{Preferential reasoning}
\label{sec:pref}
\subsection{The system {\bf P}}
\label{subsec:P}
We shall now consider a system that seems to 
occupy a central position in the hierarchy of nonmonotonic systems. 
It is strictly stronger than {\bf CL}, but assumes the existence of
disjunction in the language of formulas.
We call this system {\bf P}, for
{\em preferential}, because its semantics, described in 
section~\ref{subsec:prefmod}, are a variation on those
proposed by Y.~Shoham in \cite{Shoham:87}.
The differences (the distinction we make and he does not between 
states and worlds) are nevertheless technically important, as noticed 
above just before definition~\ref{def:asy}, and as will be shown at the end of 
section~\ref{subsec:prefmod}.
This very system has been considered by E.~Adams~\cite{Adams:75} and
proposed by J.~Pearl and H.~Geffner \cite {PearlGeff:88} to serve
as the {\em conservative core} of a nonmonotonic reasoning system.
It is the flat fragment of the system ${\cal S}$ studied by J.~Burgess
in \cite{Burgess:81}. 

\begin{definition}
\label{def:P}
The system {\bf P} consists of all the rules of {\bf C} and
the following:
\begin{equation}
\label{ru:or}
{{\alpha  \NI \gamma \ \ , \ \ \beta  \NI \gamma} \over
{\alpha \vee \beta  \NI \gamma}} \hspace {2cm} {\rm ({\bf Or})}
\end{equation}
A consequence relation that satisfies all rules of {\bf P} is
said to be {\em preferential}.
\end{definition}

The rule {\bf Or} corresponds to the axiom CA of conditional logic.
It says that any formula that is, separately,
a plausible consequence of two different formulas, should also be
a plausible consequence of their disjunction. It is a valid principle
of monotonic classical reasoning and does not imply monotonicity,
therefore we tend to accept it. Further consideration also seems to
support {\bf Or}: if we think that 
{\em if John attends the party, normally, the evening will be great} 
and also that 
{\em if Cathy attends the party, normally, the evening will be great} and
hear that at least one of Cathy or John will attend the party, 
shouldn't we be tempted to join in?
There is, though, an {\em epistemic} reading
of \mbox{$\alpha$ \NI $\beta$} that invalidates the {\bf Or} rule.
If we interpret 
\mbox{$\alpha$ \NI $\beta$} as meaning: 
{\em if all I know about the world is $\alpha$ then it is sensible for me
to suppose that $\beta$ is true}, we must reject the {\bf Or}
rule. Indeed, one may imagine a situation in which $\alpha$ expresses 
a fact that can very well be true or false but the truth value of which
is normally not known to me. If I knew $\alpha$ to be true, 
that would be quite an 
abnormal situation in which I may be willing to accept $\gamma$. If I
knew $\alpha$ to be false, similarly, it would be an exceptional situation
in which I may accept $\gamma$, but the knowledge that 
\mbox{$\alpha \vee \neg \alpha$} 
is true is essentially void and certainly does not
allow me to conclude that anything exceptional is happening. 
Notice that, in this reading, the left hand side of the symbol \NI\
involves a hidden epistemic operator (the right hand side may also do so,
but need not). 
We shall therefore defend
the {\bf Or} rule by saying that such a hidden operator should be made
explicit and the example just above only invalidates the inference:
from \mbox{${\cal K}\alpha$ \NI $\gamma$} and 
\mbox{${\cal K}\beta$ \NI $\gamma$}
infer \mbox{${\cal K}(\alpha \vee \beta)$ \NI $\gamma$}. 
But nobody would defend such an inference anyway.

The interplay between {\bf Or} and the rules of {\bf C} makes 
{\bf P} a powerful system. For example, {\bf Loop} is a derived rule
of {\bf P}. Since this result will be obvious once we have characterized
preferential relations semantically, we shall leave a proof-theoretic
derivation of {\bf Loop} in {\bf P} for the reader to find.

We shall now put together a number of remarks revolving around 
the rule {\bf Or}.
Our first remark is that we may derive from {\bf Or} a rule
that is similar to the {\em hard} half of the deduction theorem.
This rule was suggested in \cite{Shoham:88}. It is a very useful
rule and expresses the fact that deductions performed under strong
assumptions may be useful even if the assumptions are not known facts.

\begin{lemma}
\label{S}
In the presence of {\bf Reflexivity}, {\bf Right Weakening} 
and {\bf Left Logical Equivalence},
the rule of {\bf Or} implies the following:
\begin {equation}
{{\alpha \wedge \beta \NI \gamma} \over
{\alpha \NI \: \beta \rightarrow \gamma}}\hspace {3.8cm} {\rm ({\bf S})}
\end {equation}
{\bf S} is therefore a derived rule of {\bf P}.
\end{lemma}
\proof
Suppose 
\mbox{$\alpha \wedge \beta$ \NI $\gamma$}. 
We have 
\mbox{$\alpha \wedge \beta$ \NI $\beta \ra \gamma$},
by {\bf Right Weakening}.
But one has
\mbox{$\alpha \wedge \neg \beta$ \NI $\beta \ra \gamma$}.
One concludes
by {\bf Or} and {\bf Left Logical Equivalence}.
\QED

Our second remark is that, in the presence of {\bf S}, the rule of
{\bf Cut} is implied by {\bf And}. Therefore {\bf Reflexivity},
{\bf Left Logical Equivalence}, {\bf Right Weakening}, {\bf And}, 
{\bf Or} and {\bf Cautious Monotonicity} are an elegant equivalent
axiomatization of the system {\bf P}.

\begin{lemma}
\label{le:A.C}
In the presence of {\bf Right Weakening}, 
{\bf S} and {\bf And} imply {\bf Cut}.
\end{lemma}
\proof
Use {\bf S}, {\bf And} and {\bf Right Weakening}.
\QED

D.~Makinson~\cite{Mak:88} suggested the following rule.
It expresses the principle of proof by cases.
\begin{equation}
\label{ru:D}
{{\alpha \wedge \neg \beta \NIm \gamma \ \ , \ \ \alpha \wedge \beta \NIm
\gamma} \ \ \ \over {\alpha \NIm \gamma}} \hspace{1cm}{\rm ({\bf D})}
\end{equation}

\begin{lemma}
\label{le:D}
In the presence of {\bf Reflexivity}, {\bf Right Weakening} 
and {\bf Left Logical Equivalence}, 
\begin{enumerate}
\item {\bf Or} implies {\bf D} and 
\item {\bf D} implies {\bf Or} in the presence of {\bf And}.
\end{enumerate}
Therefore {\bf D} is a derived rule of {\bf P}.
\end{lemma}

The proof is left to the reader.

The next lemma gathers some more derived rules of the system {\bf P}.
They will be used in the proof of the representation theorem of
section~\ref{subsec:pref.char}.
The importance of these rules is mainly technical.
The reader should notice that
{\bf P} is a powerful system, in which one may build quite
sophisticated proofs.

\begin{lemma}
\label{lemma:ded}
The following are derived rules of {\bf P}:

\begin{equation}
\label{or2}
{{\ \ {\alpha \NIm \gamma} \ \ , \ \ {\beta \NIm \delta}} \ \ \ \over
{{\alpha \vee \beta}  \NIm {\gamma \vee \delta}}}
\end{equation}

\begin{equation}
\label{or:implies1}
{{\ \ {\alpha \vee \gamma} \NIm \gamma \ \ , 
\ \ \alpha \NIm \beta} \ \ \ \over
{\gamma \NIm {\alpha \rightarrow \beta}}}
\end{equation}

\begin{equation}
\label{Less:Trans}
{{\ \ {\alpha \vee \beta} \NIm \alpha \ \ , 
\ \ {\beta \vee \gamma} \NIm \beta}
\ \ \ \over
{{\alpha \vee \gamma} \NIm \alpha}}
\end{equation}

\begin{equation}
\label{new1}
{{\ \  {\alpha \vee \beta}  \NIm \alpha \ \ , 
\ \ {\beta \vee \gamma} \NIm \beta}
\ \ \ \over
{\alpha \NIm {\gamma \rightarrow \beta}}}
\end{equation}
\end{lemma}
\proof
The uses of {\bf Left Logical Equivalence} will not always be mentioned 
any more.
For (\ref{or2}), use first {\bf Right Weakening} on each of the two hypotheses 
and then {\bf Or}. This seems to be a very intuitive rule that is
often useful.

For (\ref{or:implies1}), from the second hypothesis, using {\bf Left Logical
Equivalence} we have 
\mbox{${( \alpha \vee \gamma ) \wedge \alpha}$ \NI $\beta$}.
By {\bf S} we conclude 
\mbox{${\alpha \vee \gamma}$ \NI ${\alpha \rightarrow \beta}$}.
But, using the first hypothesis and {\bf Cautious Monotonicity} one may now 
conclude. 

For (\ref{Less:Trans}), from both hypotheses and using (\ref{or2})
one concludes
\mbox{${\alpha \vee \beta \vee \gamma}$ \NI ${\alpha \vee \beta}$}.
Now, using our first hypothesis and (\ref{Or:Trans}) we see 
\mbox{${( \alpha \vee \beta ) \vee \gamma}$ \NI $\alpha$}.
Leaving this result for a moment, notice that from the first hypothesis
and \mbox{\gc \NI \gc}, using (\ref{or2}) we obtain 
\mbox{${\alpha \vee \beta \vee \gamma}$ \NI ${\alpha \vee \gamma}$}. 
Now, coming back to the result we left hanging, using 
{\bf Cautious Monotonicity}, we may conclude.

For (\ref{new1}), from the second hypothesis one has 
\mbox{${( \beta \vee \gamma ) \wedge ( \alpha \vee \beta \vee \gamma )}$ \NI 
$\beta$}.
By {\bf S}: 
\mbox{${\alpha \vee \beta \vee \gamma}$ \NI ${( \beta \vee \gamma )
\rightarrow \beta}$}.
By {\bf Right Weakening}, one may then obtain
\mbox{${\alpha \vee \beta \vee \gamma}$ \NI ${\gamma \rightarrow \beta}$}
But from the two hypotheses, using (\ref{or2}), one obtains:
\mbox{${\alpha \vee \beta \vee \gamma}$ \NI ${\alpha \vee \beta}$}.
Using {\bf Cautious Monotonicity} on those last two results, we obtain:
\mbox{${\alpha \vee \beta}$ \NI ${\gamma \rightarrow \beta}$}.
Using the first hypothesis and {\bf Cautious Monotonicity} one concludes.
\QED

\subsection{Preferential Models}
\label{subsec:prefmod}
We may now describe our version of preferential models.
Preferential models are cumulative ordered models in which states are
labeled by single worlds (and not sets of worlds).
The reasoner has, then, essentially, a preference over worlds
(except that the same world may label different states). 
We may now define the family of models we are interested in.

\begin{definition}
\label{def:prefmod}
A {\em preferential} model $W$ is a triple 
\mbox{$\langle \: S, l, \prec \rangle$}
where $S$ is a set, the elements of which will be called states, 
\mbox{$l:S \mapsto {\cal U}$} assigns a world to each state 
and $\prec$ is a strict partial order on $S$ (i.e. an
irreflexive, transitive relation), satisfying the 
{\bf smoothness condition} of definition~\ref{def:smoocond}.
\end{definition}

Notice that, for a preferential model,
\mbox{$s \EM \alpha$} iff \mbox{$l ( s ) \models \alpha$}.
The smoothness condition, here, as explained in section~\ref{subsec:cum.ord},
is only a technical condition needed to deal with
infinite sets of formulas, it is always satisfied in any preferential 
model in which 
$S$ is finite, or in which $\prec$ is well-founded 
(i.e. no infinite descending chains).
The requirement that the relation $\prec$ be a strict partial order has been
introduced only because such models are nicer and the smoothness condition
is easier to check on those models, but the soundness result of lemma
\ref{soundness:non} is true for the larger family of models, where $\prec$ is
just any binary relation. 
In such a case, obviously, the smoothness condition cannot be dropped even
for finite models.
The completeness
result of theorem \ref{Com:pref} holds obviously, too, for the larger family,
but is less interesting.
Preferential models, since they are cumulative models, define
consequence relations as in definition~\ref{def:cumcons}.

Y.~Shoham, in \cite{Shoham:87}, proposed a more restricted notion
of preferential models. He required the set of states $S$ to be a subset
of the universe \cU\ and the labeling function $l$ to be the identity.
He also required the relation $\prec$ to be a well-order.
Any one of those two requirements would make the representation
theorem incorrect. The second point is treated in \cite{LMTR:88}.
For the first point, we leave it as an exercise to the reader 
to show that the following model has no equivalent model in which
no label appears twice. Let $L$ be the propositional calculus on
two variables $p$ and $q$. Let $S$ have four states: $s_0 \prec s_2$
and $s_1 \prec s_3$. Let $s_0$ satisfy $p$ and $\neg q$, $s_1$ satisfy
$\neg p$ and $\neg q$ and $s_2$ and $s_3$ both satisfy $p$ and $q$.

\subsection{Characterization of preferential consequence relations}
\label{subsec:pref.char}
Our first lemma is obvious. It does not hold in cumulative models
and should be contradistincted with lemma~\ref{remark1}.

\begin{lemma}
\label{remark2}
Let \mbox{$W = \langle \: S , l , \prec \rangle$} be a preferential model.
For any \mbox{$\alpha , \beta \in L$},
\mbox{${\widehat{\alpha \vee \beta}} = 
{\widehat{\alpha} \cup \widehat{\beta}}$}.
\end{lemma}

\begin{lemma}[Soundness]
\label{soundness:non}
For any preferential model $W$, the consequence relation \NIW\ it
defines is a preferential relation, i.e. all the rules
of the system {\bf P} are satisfied by the relations defined by 
preferential models.
\end{lemma}
\proof
Indeed, as we remarked above, the fact that $\prec$
is a partial order is not used at all.
Since a preferential model is a cumulative model, in light of
lemma~\ref{Cu:Sou}, we only need to check the validity of {\bf Or}.
Suppose a preferential model \mbox{$W = \langle S, l, \prec \rangle$}
and \mbox{$\alpha , \beta , \gamma \in L$} are given.
Suppose that \mbox{$\alpha$ \NIW $\gamma$} and 
\mbox{$\beta$ \NIW $\gamma$}.
Any state minimal in \mbox{$\widehat{\alpha \vee \beta}$} is, 
by lemma~\ref{remark2},
minimal in the set \mbox{$\widehat\alpha \cup \widehat\beta$}, 
and therefore minimal 
in any of the subsets it belongs to.
\QED

We shall now begin the proof of the representation theorem.
Let us, first, define a relation among formulas, that will turn out to be a 
pre-ordering whenever the relation \NI\ satisfies {\bf P}.

\begin{definition}
\label{def:order}
We say that $\alpha$ is not less {\em ordinary} than $\beta$ and write 
\mbox{$\alpha \leq \beta$} iff \mbox{$\alpha \vee \beta$ \NI $\alpha$}.
\end{definition}

Indeed, if we would conclude that $\alpha$ is true on the basis
that either $\alpha$ or $\beta$ is true, this means that the former is not 
more out of the ordinary than the latter.
Notice that, if \NI\ satisfies {\bf Reflexivity} and {\bf Left Logical 
Equivalence}, then for any \mbox{$\alpha , \beta \in L$},
\mbox{$\alpha \vee \beta \leq \alpha$}.

\begin{lemma}
\label{lemma:order}
If the relation \NI\ is preferential, the relation 
$\leq$ is reflexive and transitive.
\end{lemma}
\proof
Reflexivity follows from {\bf Left Logical Equivalence} and {\bf Reflexivity}.
Transitivity follows from (\ref{Less:Trans}) of lemma~\ref{lemma:ded}.
\QED

From now on, and until theorem~\ref{Com:pref}, we shall suppose that 
the relation \NI\ is preferential.
\begin{lemma}
\label{le:upnorm}
If \mbox{$\alpha \leq \beta$}
and $m$ is a normal world for $\alpha$ that satisfies $\beta$, then $m$
is a normal world for $\beta$.
\end{lemma}
\proof
Suppose \mbox{$\beta$ \NI $\delta$}. 
By (\ref{or:implies1}) of lemma~\ref{lemma:ded}, we have
\mbox{$\alpha$ \NI $\beta \ra \delta$}. If $m$ is normal for $\alpha$
it must satisfy \mbox{$\beta \ra \delta$}, and since it satisfies $\beta$,
it must satisfy $\delta$.
\QED
\begin{lemma}
\label{lemma:chain}
If \mbox{$\alpha \leq \beta \leq \gamma$} and $m$ is a normal 
world for $\alpha$
that satisfies $\gamma$ then it is a normal world for $\beta$.
\end{lemma}
\proof
By lemma~\ref{le:upnorm}, it is enough to show that $m$ satisfies $\beta$.
By (\ref{new1}) of lemma~\ref{lemma:ded} we have 
\mbox{$\alpha$ \NI $\gamma \ra \beta$}, 
but $m$ is a normal world for $\alpha$ that satisfies $\gamma$, 
therefore it must satisfy $\beta$.
\QED

We may now describe the preferential model we need for the representation 
result.
Remember that we start from any preferential relation \NI.
We then consider the following model:
\mbox{$W \eqdef \langle \: S , l , \prec \rangle$}
where
\begin{enumerate}
\item
$S \eqdef \{ {< m , \alpha >} \mid {m \mbox{ is a normal world for } \alpha} 
\}$,
\item
\mbox{$l ( < m , \alpha > ) = m$}
and 
\item
\mbox{$<m, \alpha> \prec <n,\beta>$} iff 
\mbox{$\alpha \le \beta$} and 
\mbox{$m \not \models \beta$}.
\end{enumerate}

The first thing we want to show is that $W$ is a preferential model,
i.e. that $\prec$ is a strict partial order and that $W$ satisfies the
smoothness condition.
We shall then show that the relation \NIW\ is exactly \NI.

\begin{lemma}
\label{lemma:strict} 
The relation $\prec$ is a strict partial order, i.e. it is irreflexive 
and transitive.
\end{lemma}
\proof
The relation $\prec$ is irreflexive since 
\mbox{$<m, \alpha> \prec <m, \alpha>$}
would imply \mbox{$m \not \models \alpha$}, 
but $m$ is a normal world for $\alpha$,
and since \mbox{$\alpha$ \NI $\alpha$} by {\bf Reflexivity}, it satisfies
$\alpha$.
It is left to show that $\prec$ is transitive.
Suppose \mbox{$<m_0,\alpha_0> \prec <m_1 , \alpha_1>$} and
\mbox{$<m_1 , \alpha_1> \prec <m_2, \alpha_2>$}.
By the definition of $\prec$ we have \mbox{$\alpha_0 \leq \alpha_1$} and 
\mbox{$\alpha_1 \leq \alpha_2$}. From this we may conclude two things.
First, by lemma~\ref{lemma:order} we conclude 
\mbox{$\alpha_0 \leq \alpha_2$}.
Secondly, since $m_0$ is a normal world for $\alpha_0$ that does not satisfy
$\alpha_1$, we may conclude by lemma~\ref{lemma:chain} that it does not satisfy
$\alpha_2$.
\QED

We are now going to characterize all minimal states in sets of the form
$\widehat{\alpha}$.
\begin{lemma}
\label{minimal:or}
In the model $W$, \mbox{$ < m , \beta > $} 
is minimal in $\widehat{\alpha}$ iff 
\mbox{$m \models \alpha$} and $\beta \leq \alpha$.
\end{lemma}
\proof
For the {\em if} part,
suppose \mbox{$m \models \alpha$} and \mbox{$\beta \leq \alpha$}.
Clearly \mbox{$m \in \widehat{\alpha}$}.
Suppose now that \mbox{$< n , \gamma > \prec < m , \beta >$} and 
\mbox{$n \models \alpha$}. 
We would have \mbox{$\gamma \leq \beta \leq \alpha$},
$n$ normal for $\gamma$, and
\mbox{$n \not \models \beta$} and \mbox{$m \models \alpha$}.
This stands in contradiction with lemma~\ref{lemma:chain}.

For the {\em only if} part, suppose
\mbox{$<m,\beta>$} is minimal in $\widehat\alpha$. Clearly 
\mbox{$m \models \alpha$}.
Suppose $n$ is a normal world for \mbox{$\alpha \vee \beta$} that does not
satisfy $\beta$ 
(it is not claimed that such a normal world exists). 
Since \mbox{$\alpha \vee \beta \leq \alpha$}, we must have
\mbox{$<n,\alpha \vee \beta> \prec <m,\beta>$}. 
But $n$ is a normal world
for \mbox{$\alpha \vee \beta$} that does not satisfy $\beta$ and therefore 
must satisfy $\alpha$.
This stands in contradiction with the minimality of
\mbox{$<m,\beta>$} in $\widehat\alpha$.
We conclude that every normal world for \mbox{$\alpha \vee \beta$}
satisfies $\beta$. By lemma~\ref{norm:mod}, 
\mbox{$\alpha \vee \beta$ \NI $\beta$}.
\QED

We shall now prove that $W$ satisfies the smoothness condition.
\begin{lemma}
\label{lemma:smoo}
For any $\alpha \in L$, $\widehat \alpha$ is smooth.
\end{lemma}
\proof
Suppose \mbox{$<m, \beta>\in \widehat \alpha$}, i.e., 
\mbox{$m \models \alpha$}.
If \mbox{$\beta \leq \alpha$} then, by lemma~\ref{minimal:or} 
\mbox{$<m,\beta>$}
is minimal in $\widehat\alpha$.
On the other hand, if \mbox{$\alpha \vee \beta$ \notNI $\beta$}
then by lemma~\ref{norm:mod} 
there is a normal world $n$ for \mbox{$\alpha \vee \beta$}
such that \mbox{$n \not \models \beta$}.
But \mbox{$\alpha \vee \beta \leq \beta$} and
therefore \mbox{$< n , \alpha \vee \beta > \prec < m , \beta >$}. 
But, \mbox{$n\models \alpha \vee \beta$} and 
\mbox{$n \not \models \beta$} therefore 
\mbox{$n \models \alpha$}.
Since \mbox{${\alpha \vee \beta} \leq \alpha$},
Lemma~\ref{minimal:or} enables us to conclude that
\mbox{$< n , \alpha \vee \beta >$} is minimal in
$\widehat{\alpha}$.
\QED

We have shown that $W$ is a preferential model. We shall now show that 
\NIW\ is exactly the relation \NI\ we started from.
\begin{lemma}
\label{le:1}
If \mbox{$\alpha$ \NI $\beta$}, then \mbox{$\alpha$ \NIW $\beta$}.
\end{lemma}
\proof
We must show that all minimal states of $\widehat{\alpha}$ satisfy $\beta$.
Suppose \mbox{$<m,\gamma>$} is minimal in $\widehat \alpha$.
Then $m$ is a normal world for $\gamma$ that satisfies $\alpha$.
By lemma~\ref{minimal:or}, 
\mbox{$\gamma \le \alpha$} and therefore, by lemma~\ref{le:upnorm},
$m$ is a normal world for $\alpha$.
\QED

\begin{lemma}
\label{le:2}
If \mbox{$\alpha$ \NIW $\beta$}, then \mbox{$\alpha$ \NI $\beta$}.
\end{lemma}
\proof
It follows from the definition of the relation $\prec$ 
(lemma~\ref{minimal:or} could also be used, but is not really necessary here)
that, 
given any normal world $m$ for $\alpha$,
\mbox{$< m , \alpha >$} is minimal in $\widehat{\alpha}$.
If \mbox{$\alpha \NIW \beta$}, $\beta$ is satisfied by all normal
worlds for $\alpha$, and we may conclude by lemma~\ref{norm:mod}.
\QED

We may now state the main result of this section.
\begin{theorem}[Representation theorem for preferential relations]
\label{Com:pref}
A consequence relation is a preferential consequence relation iff it
is defined by some preferential model.
\end{theorem}
\proof
The {\em if} part is Lemma~\ref{soundness:non}.
For the {\em only if} part, let \NI\ be any consequence relation satisfying
the rules above and let
$W$ be defined as above.
Lemmas~\ref{lemma:strict} and \ref{lemma:smoo} show 
that $W$ is a preferential model.
Lemmas~\ref{le:1} and \ref{le:2} show that it defines an consequence relation
that is exactly \NI.
\QED

As in the cumulative and cumulative ordered cases we may study 
the notion of preferential entailment and obtain results similar
to Corollaries~\ref{log:imp3}, \ref{co:cum2} and \ref{comp:cum}.

\subsection{Some rules that cannot be derived in {\bf P}}

Is {\bf P} a reasonable system for nonmonotonic reasoning?
We think a good reasoning system should validate all the rules of
{\bf P}.
Notice that all the rules we have considered so far are of the
form: from the presence of certain assertions in the consequence
relation, deduce the presence of some other assertion.
After careful consideration of many other rules of this form, 
we may say we have good reason to think that there are no rules
of this type that should be added.
Certain principles of reasoning that seem appealing, though, fail to be 
validated by certain preferential consequence relations. This means, 
in our sense, that many agents that reason in a way that is fully 
consistent with all the rules of {\bf P}, nevertheless behave irrationally.
We shall show that circumscription does not, in general, satisfy even
the weakest of the principles we shall present.
The reader will notice that the form of these principles is different
from that of all the rules previously discussed:
from the {\em absence} of certain assertions in the relation,
we deduce the {\em absence} of some other assertion.

\begin{equation}
\label{Neg:mon}
{{\alpha \wedge \gamma \notNIm \beta \ , 
\ \alpha \wedge \neg \gamma \notNIm \beta} 
\over
{\alpha \notNIm \beta}} \hspace {1.6cm} 
{\rm ({\bf Negation \ Rationality}) }
\end{equation}

\begin{equation}
\label{Or:mon}
{{\alpha \notNIm \gamma \ \ , \ \ \beta \notNIm \gamma}
\over {\alpha \vee \beta \notNIm \gamma}} \hspace {3cm}
{\rm ({\bf Disjunctive \ Rationality}) }
\end{equation}

\begin{equation}
\label{Rat:mon}
{{\alpha \wedge \beta \notNIm \gamma \ \ , \ \ \alpha \notNIm \neg\beta} 
\over
{\alpha \notNIm \gamma}} \hspace {2cm}
{\rm ({\bf Rational \ Monotonicity}) }
\end{equation}

Each one of those rules is implied by {\bf Monotonicity} and 
therefore expresses some kind of restricted monotonicity.
Any rational reasoner should, in our opinion, support them, and we shall,
now, explain and justify them.
The rule of {\bf Negation Rationality} says 
that inferences are not made solely on the basis of ignorance.
If we accept that \gb\ is a plausible consequence of \ga, we must either
accept that it is a plausible consequence of 
\mbox{$\alpha \wedge \gamma$}
or accept that it is a plausible consequence of 
\mbox{$\alpha \wedge \neg \gamma$}.
Indeed, suppose we hold that {\em normally, the party should be great},
but that we do not hold that {\em even if Peter comes to the party, it will
be great}, i.e. we seriously doubt the party could stand Peter's presence.
It seems we could not possibly hold that we also seriously doubt that
the party could stand Peter's absence. If we do not expect the party to
be great if Peter is there and do not expect it to be great if Peter is not
there, how could we expect it to be great? After all, either Peter is going to
be there or he is not.
It is, though, easy to find examples of preferential models that define
consequence relations that do not satisfy {\bf Negation Rationality}.

We shall even show, now, that circumscriptive reasoning does
not always obeys {\bf Negation Rationality}.
Suppose our language has two unary predicate symbols {\em special}
and {\em beautiful}, and one individual constant $a$.
We know that, {\em normally an object is not special}, i.e. we
circumscribe by minimizing the extension of {\em special}, keeping 
{\em beautiful} constant.
Take \ga\ to be {\bf true} and \gb\ to be \mbox{$\neg {\em special}(a)$}.
Indeed, without any information, we shall suppose that $a$ is not special.
But take \gc\ to be 
\mbox{${\em beautiful} ( a ) \leftrightarrow {\em special} ( a )$}.
If we had the information that $a$ is beautiful if and only if it is special,
we could not conclude that $a$ is not special anymore, since it could
well be beautiful, i.e. there are two {\em minimal} models that must be
considered: the first one with $a$ neither beautiful nor special and
the second one with $a$ beautiful and special.
On the other hand, had we had the informatiom that either $a$ is
beautiful or it is special but not both, we could not have concluded
that it is not special either, since it could well not be beautiful.
It seems that circumscription may lead to unexpected conclusions.
The example presented here is a simplification, due to M.~Ginsberg, of an
example due to the second author.
If we try to understand where circumscription differs from intuitive
reasoning, we probably will have to say that, even with the knowledge
that $a$ is special if and only if it is beautiful, we would have kept
the expectation that it is not special, and therefore gained the expectation 
that it is not beautiful. Similarly, with the knowledge that $a$ is
either special or beautiful but not both, we would have kept the
expectation that it is not special and therefore formed 
the expectation that it is beautiful.

The rule of {\bf Disjunctive Rationality} says
that inferences made from a disjunction of propositions
must be supported by at least one of the component propositions.
Again, this seems like a reasonable requirement.
If we do not hold that {\em if Peter comes to the party, it will be great}
and do not hold that {\em if Cathy comes to the party, it will be great},
how could we hold that {\em if at least one of Peter or Cathy comes, the
party will be great}? In this example, the reader may prefer to read
{\em even if} instead of {\em if}, but the conclusion stands anyway.
It is easy to see that {\bf Disjunctive Rationality} implies
{\bf Negation Rationality}. The second author recently showed that
{\bf Disjunctive Rationality} is strictly stronger than {\bf Negation
Rationality}.

The rule of {\bf Rational Monotonicity} is similar to 
the axiom CV of conditional logic.
It expresses the fact that only additional information the 
negation of which was expected
should force us to withdraw plausible conclusions previously drawn.
It is an important tool in minimizing the updating we have to do
when learning new information.
Suppose we hold that {\em normally, the party will be great} but
do not hold that {\em even if Peter comes, the party will be great},
i.e. we think Peter's presence could well spoil the party, shouldn't
we hold that {\em normally, Peter will not come to the party}?
One easily shows that, in the presence of the rules of {\bf C},
{\bf Rational Monotonicity} implies {\bf Disjunctive Rationality}.
D.~Makinson proved
that {\bf Rational Monotonicity} is strictly stronger than
{\bf Disjunctive Rationality} and 
conjectured a model-theoretic characterization of preferential 
relations that satisfy {\bf Rational Monotonicity}.
The second author proved the corresponding representation 
result in the case the language $L$ is finite. The third author 
lifted the restriction on $L$.
These results will appear in a separate paper.

\subsection{Examples: diamonds and triangles}
\label{subsec:exa}
We shall now show what preferential reasoning may provide in the setting
of two toy situations that have become classics in the literature.
First the so-called {\em Nixon diamond}.
Suppose our knowledge base \bK\ contains the four assertions that follow.
The reader may read {\em teen-ager} for $t$, {\em poor} for $p$, 
{\em student} for $s$ and {\em employed} for $e$.
\begin{enumerate}
\item $t$ \NI $p$
\item $t$ \NI $s$
\item $p$ \NI $e$
\item $s$ \NI $\neg e$
\end{enumerate}
It is easy to see, by describing suitable preferential models, that 
no assertion that would look like some kind of contradiction is 
preferentially entailed
by \bK. In particular neither 
\mbox{$t$ \NI $e$}, nor \mbox{$t$ \NI $\neg e$} is preferentially
entailed by \bK.
We cannot conclude, from the information given above, that teen-agers are
normally employed, neither can we conclude that they generally are not
employed.
This seems much preferable than the consideration of multiple extensions.
This weakness of the system {\bf P} seems to be exactly what we want.
Nevertheless, preferential reasoning allows for some quite 
subtle conclusions. For example the following assertions are preferentially
entailed by \bK:
\mbox{{\bf true} \NI $\neg t$} ({\em normally, people are not teen-agers}),
\mbox{{\bf true} \NI $\neg ( p \wedge s )$} ({\em normally, people are not
poor students}). The following assertions are not preferentially
entailed:
\mbox{$s$ \NI $\neg p$} ({\em students, normally are not poor}),
or \mbox{$p$ \NI $\neg s$} ({\em poor persons are normally not students}), 
and we feel indeed that there is not enough 
information in \bK\ to justify them.
An example of an assertion that is not preferentially entailed 
by \bK\ but we think should follow from \bK\ is:
\mbox{$a \wedge p$ \NI $e$}, since $a$ is not mentioned in \bK. 
The reader may consult \cite{Leh:89} for a possible solution.

A second classical example is the {\em penguin triangle}.
Suppose our knowledge base \bK\ contains the three assertions that follow.
The reader may read {\em penguin} for $p$, {\em flies} for $f$, 
and {\em bird} for $b$.
\begin{enumerate}
\item $p$ \NI $b$
\item $p$ \NI $\neg f$
\item $b$ \NI $f$
\end{enumerate}

It is easy to see, by describing suitable preferential models, that 
no assertion that would lead to some kind of contradiction is 
preferentially entailed
by \bK. In particular \mbox{$p$ \NI $f$} is not preferentially
entailed by \bK.
On the other hand, the following assertions are preferentially entailed
by \bK\ and we leave it to 
the reader to show that they are satisfied by all preferential models 
that satisfy \bK:
\begin{enumerate}
\item \mbox{$p \wedge b$ \NI $\neg f$}
\item \mbox{$f$ \NI $\neg p$}
\item \mbox{$b$ \NI $\neg p$}
\item \mbox{$b \vee p$ \NI $f$}
\item \mbox{$b \vee p$ \NI $\neg p$}
\end{enumerate}

The reader should remark that no {\em multiple extension} problem arises
here and that preferential reasoning correctly chooses the most specific
information and in effect pre-empts the application of a less specific
default.

\subsection{Horn assertions}
\label{subsec:horn}
In this section we shall show that, if we consider only assertions of a 
restricted type (i.e. Horn assertions), then the system {\bf P} is no
stronger than {\bf CL}. For this result we shall need the full strength
of theorem~\ref{rep:PC}. To keep notations
simple, let us suppose $L$ is a propositional language.

\begin{definition}
\label{def:horn}
An assertion \ab\ will be called a Horn assertion iff the antecedent
\ga\ is a conjunction of zero or more propositional variables and
the consequent \gb\ is either a single propositional variable or 
the formula {\bf false}.
\end{definition}

The crucial remark is the following.

\begin{lemma}
\label{le:ordcum.constr}
If $W$ is a cumulative ordered model, there is a preferential model
$V$ such that \NIW\ and \NIV\ coincide as far as Horn assertions are
concerned.
\end{lemma}
\proof
Let $W$ be the model \mbox{$\langle \: S , l , \prec \rangle$}.
We shall define $V$ to be the model
\mbox{$\langle \: S , l' , \prec \rangle$},
where $l'$ is defined in the following way.
For any \mbox{$s \in S$} and for any propositional variable $p$, 
\mbox{$l'(s) \models p$} iff for every \mbox{$u \in l(s)$},
\mbox{$u \models p$}, in other words iff \mbox{$s \EM p$} in $W$.
It is clear that, if \ga\ is a conjunction of propositional variables
then the sets $\widehat\alpha$ in $W$ and $V$ coincide. Therefore,
if $W$ satisfies the smoothness condition, so does $V$ and \NIW\ and 
\NIV\ agree on Horn formulas.
\QED

\begin{theorem}
\label{the:PLOOP}
Let \bK\ be a knowledge base containing only Horn assertions, 
and \cA\ a Horn assertion.
If the assertion \cA\ may be derived from \bK\ in the system {\bf P}, 
then it may be derived from \bK\ in the system {\bf CL}.
\end{theorem}
\proof
Suppose \cA\ cannot be derived in {\bf CL}. By the representation 
theorem~\ref{rep:PC}, there is a cumulative ordered model $W$
that satisfies all the assertions of \bK, but does not satisfy
\cA. By lemma~\ref{le:ordcum.constr}, there is a preferential
model $V$ that satisfies \bK, but does not satisfy \cA.
We conclude, by the soundness part of theorem~\ref{Com:pref},
that \cA\ cannot be derived in {\bf P}.
\QED

\section{Cumulative monotonic reasoning}
\label{sec:mon.cum}
\subsection{The system {\bf CM}}
\label{subsec:CM}
In section~\ref{subsec:m}, three rules were shown equivalent in the
presence of the rules of {\bf C}. We shall now study the system 
obtained by adding those rules (or one of them) to the system {\bf C}.
One obtains a system that is strictly stronger than {\bf CL}, but
incomparable with {\bf P}. It is corresponds to some natural
family of models.

\begin{definition}
The system {\bf CM} contains all the rules of {\bf C} and the rule
of {\bf Monotonicity}, defined in (\ref{ru:mon}).
A consequence relation that satisfies all the rules of {\bf CM}
is said to be {\em cumulative monotonic}.
\end{definition}

In fact, {\bf Left Logical Equivalence} and {\bf Cautious Monotonicity}
are now redundant, since they follow from {\bf Monotonicity}.
From lemma~\ref{le:dermon}, one sees that {\bf EHD} and {\bf Transitivity}
are derived rules of {\bf CM}. It is obvious that {\bf Loop} is also
a derived rule of {\bf CM} (by {\bf Transitivity}).
It is not difficult to find preferential models that do not satisfy
{\bf Monotonicity} and we conclude that {\bf CM} is strictly stronger than
{\bf CL} and not weaker than {\bf P}.

\subsection{Simple cumulative models}
\label{subsec:simcum.mod}
\begin{definition}
A cumulative model will be called a {\em simple} cumulative model iff
the binary relation $\prec$ on its states is empty.
\end{definition}

A simple cumulative model is a cumulative ordered model.
The smoothness condition is always satisfied in such a model.
It is very easy to see that the consequence relation defined by 
any simple cumulative model satisfies {\bf Monotonicity}.
It is not difficult to find simple cumulative models that do not
satisfy certain instances of the {\bf Or} rule.
We conclude that {\bf P} and {\bf CM} are incomparable.
It is also easy to find such models that do not satisfy certain 
instances of {\bf Contraposition}.

\subsection{Characterization of monotonic cumulative consequence relations}

\begin{theorem}[Representation theorem for cumulative monotonic relations]
\label{char:MC}
A consequence relation is cumulative monotonic iff it is defined by some
simple cumulative model.
\end{theorem}
\proof
It has been noticed above that the {\em if} part is trivial.
For the {\em only if} part, suppose \NI\ is a consequence relation that
satisfies the rules of {\bf CM}.
Let \mbox{$W \eqdef \langle A , l , \emptyset \rangle$}, where
\mbox{$A \subseteq L$} is the set of all formulas \ga\ such that
\mbox{\ga \notNI {\bf false}} and 
$l \eqdef$ \mbox{$\{ m \mid m $\mbox{ is a normal world for }$ \alpha \}$}.
Lemma~\ref{norm:mod} implies that all labels are non-empty.
By lemma~\ref{norm:mod}, for any formula $\alpha$, 
\mbox{$\widehat{\alpha} = \{ \beta \mid \beta \NI \alpha \}$}.
Since all states of $\widehat{\alpha}$ are minimal in $\widehat{\alpha}$,
we see that \mbox{$\alpha$ \NIW $\beta$} iff for all $\gamma$ such that 
\mbox{$\gamma$ \NI $\alpha$} and all normal worlds $m$ for $\gamma$, 
\mbox{$m \models \beta$}. 
By lemma~\ref{norm:mod} this last condition is equivalent
to \mbox{$\gamma$ \NI $\beta$} and we have:
\mbox{$\alpha$ \NIW $\beta$} iff for any $\gamma$,
\mbox{$\gamma$ \NI $\alpha \Ra \gamma \NI \beta$}.
Suppose \mbox{$\alpha$ \NI $\beta$}, take any $\gamma$ such that 
\mbox{$\gamma$ \NI $\alpha$},
we have by {\bf Transitivity}, a derived rule of {\bf CM},
\mbox{$\gamma$ \NI $\beta$}. Therefore 
\mbox{$\alpha$ \NIW $\beta$}.
Suppose now that \mbox{$\alpha$ \NIW $\beta$}, then, by taking 
\mbox{$\gamma = \alpha$}
one sees that \mbox{$\alpha$ \NI $\beta$}.
\QED

As in the cumulative, cumulative ordered and preferential cases,
one may study the notion of entailment 
yielded by simple cumulative models and obtain
results similar to Corollaries~\ref{log:imp3}, \ref{co:cum2} 
and \ref{comp:cum}.

\section{Monotonic reasoning}
\label{sec:M}
\subsection{The system {\bf M}}
The results presented in this section are probably folklore.
They are presented here for completeness' sake.
\begin{definition}
\label{def:M}
The system {\bf M} consists of all the rules of {\bf C} and
the rule of {\bf Contraposition}.
A consequence relation that satisfies all the rules of {\bf M}
is said to be {\em monotonic}.
\end{definition}

Lemma~\ref{le:dermon2} and the results to come will show that 
the system {\bf M} is strictly stronger than {\bf P} and {\bf CM}.

\begin{lemma}
\label{le:derM}
The rule {\bf Or} is a derived rule of {\bf M}.
\end{lemma}
\proof
Use {\bf Contraposition} twice, then {\bf And} and finally
{\bf Contraposition}.
\QED

\begin{lemma}
\label{le:equivM}
A consequence relation is monotonic iff it satisfies {\bf Reflexivity},
{\bf Right Weakening}, {\bf Monotonicity}, {\bf And} and {\bf Or}.
\end{lemma}
\proof
The {\em only if} part follows from lemmas~\ref{le:dercum},
\ref{le:dermon2} and \ref{le:derM}.
For the {\em if} part, notice, first, that {\bf Left Logical Equivalence}
and {\bf Cautious Monotonicity} are special cases of {\bf Monotonicity}.
The remark preceding lemma~\ref{le:A.C} shows that all rules of
{\bf P} may be derived from the rules above.
We must now show that {\bf Contraposition} may be derived from the rules
of {\bf P} and {\bf Monotonicity}.
Suppose \ab. By {\bf S}, one has 
\mbox{{\bf true} \NI $\alpha \ra \beta$}.
By {\bf Right Weakening}, we conclude
\mbox{{\bf true} \NI $\neg \beta \ra \neg \alpha$}.
By {\bf Monotonicity}, we have
\mbox{$\neg \beta$ \NI $\neg \beta \ra \neg \alpha$}.
We conclude by {\bf Reflexivity} and {\bf MPC}.
\QED

\subsection{Simple preferential models}
\label{subsec:spm}
The account of monotonic reasoning that we propose is essentially the 
following.
The agent has in mind a set of possible worlds $V$: this is the set of worlds
the agent thinks are possible in practice.
This set $V$ is a subset of the set ${\cal U}$ of all logically 
possible worlds.
The agent is willing to conclude $\beta$ from $\alpha$ if all worlds of $V$
that satisfy $\alpha$ also satisfy $\beta$.

\begin{definition}
A {\em simple} preferential model is a preferential model in which
the binary relation $\prec$ is empty.
\end{definition}

A simple preferential model is a simple cumulative model in which
the labeling function $l$ labels each state with a single world.
Since repeated labels are obviously useless we could, as well, have
considered a model to be a subset of \cU.

\subsection{Characterization of monotonic consequence relations}
\label{subsec:charmon}

\begin{theorem}[Representation theorem for monotonic relations]
\label{completeness:mon}
A consequence relation is monotonic iff it is defined by some simple 
preferential model.
\end{theorem}
\proof
The proof of the {\em if} part is trivial.
For the {\em only if} part we shall build a simple preferential model 
for any given monotonic consequence relation
\NI.
Let $V \eqdef \{ m \in {\cal U} \: | \: \forall \alpha$,$\beta \in L$, if 
\mbox{$\alpha$ \NI  $\beta$} then 
\mbox{$m \models \alpha \rightarrow \beta \: \}$}
and let \mbox{$W \eqdef \langle \: V, l \rangle$} where $l$ is 
the identity function.
So, \mbox{$m \EM \alpha$} iff \mbox{$m \models \alpha$}.

We shall prove that \mbox{$\alpha$ \NI $\beta$} iff 
\mbox{$\alpha$ \NIW $\beta$}. 
If \mbox{$\alpha$ \NI $\beta$} then by the construction of $V$,
\mbox{$\alpha$ \NIW $\beta$}.
Suppose now that \mbox{$\alpha$ \notNI $\beta$}, 
we shall show that there is a world 
\mbox{$m\in V$} that does not satisfy \mbox{$\alpha \rightarrow \beta$}.
Let \mbox{$\Gamma_{0} \eqdef \{ \neg \beta\} \cup 
\{ \delta |\: \alpha \NIm \delta \}$}.
Since \mbox{$\alpha$ \notNI $\beta$}, $\Gamma_0$ is satisfiable
(the full proof is given in a more general
case in lemma~\ref{norm:mod}).
Let $m$ be a world 
that satisfies $\Gamma_0$.
We shall prove that \mbox{$\forall \varphi  , \psi \in L$} 
if \mbox{$\varphi$ \NI $\psi$} then
\mbox{$m \models \varphi \rightarrow \psi$}.
If \mbox{$\varphi$ \NI $\psi$} then 
\mbox{{\bf true} \NI $\varphi \rightarrow \psi$} by {\bf S} and
\mbox{$\alpha$ \NI $\varphi \rightarrow \psi$} by {\bf Monotonicity}.
Therefore, \mbox{$\varphi \rightarrow \psi \in \Gamma_0$} 
by the definition of $\Gamma_0$,
and \mbox{$m \models \varphi \rightarrow \psi$}.
We conclude that \mbox{$m \in V$} and clearly \mbox{$m \EM \alpha$} but 
\mbox{$m \: \notEM \beta$}.
\QED

It will now be shown that all the constructions and results described 
above relativize without problems to a given set of conditional assertions.

\begin{corollary}
\label{log:imp1}
Let {\bf K} be a set of conditional assertions, and $\alpha , \beta \in L$.
Let $\Delta \eqdef$ \mbox{$\{ \gamma \rightarrow \delta \mid \gamma \NIm 
\delta \in {\bf K} \}$} and let $W$ be the monotonic model
\mbox{$\langle \: {\cal U}_\Delta , l \: \rangle$}, 
where $l$ is the identity function.
The notation ${\cal U}_\Delta$ has been defined in section~\ref{subsec:lan}.
The following conditions 
are equivalent. If they are satisfied we shall say that {\bf K} 
monotonically entails \ab.
\begin{enumerate}
\item 
\label{one}
for all monotonic models $V$ such that \NIV\ contains {\bf K}, 
\mbox{$\alpha$ \NIV $\beta$} 
\item 
\label{two}
\mbox{$\alpha$ \NIW $\beta$}
\item 
\label{three}
\mbox{$\alpha \NI \beta$}  has a proof from {\bf K} in 
the system {\bf M}.
\item 
\label{four}
\mbox{$\alpha \rightarrow \beta$} follows logically 
(with respect to ${\cal U}$) 
from the formulas of $\Delta$.
\end{enumerate}
\end{corollary}
\proof
We shall first show the equivalence of \ref{one} and \ref{two}.
The relation defined in \ref{one} is the intersection 
of all those monotonic consequence
relations that contain {\bf K}. 
If $V$ is any monotonic model such that \NIV\ contains {\bf K} 
then the 
labels of its states must be in ${\cal U}_\Delta$ (as defined in \ref{two}) 
and therefore \mbox{$\NIV$} contains \NIW. 
But \NIW\ contains {\bf K} and is one of the
relations considered in \ref{one}.
To see the equivalence of \ref{one} and \ref{three}, notice that
the relation defined in \ref{three} is 
the intersection of all those monotonic relations that
contain {\bf K}.
Theorem~\ref{completeness:mon} implies that \ref{one} and 
\ref{three} define the same relation.
The equivalence between \ref{two} and \ref{four} is immediate.
\QED

From the equivalence of conditions \ref{one} and \ref{three} 
one easily proves the following
compactness result:
\begin{corollary}[compactness]
\label{comp:mon}
{\bf K} monotonically entails
\mbox{$\alpha$ \NI $\beta$} iff a finite subset of {\bf K} does.
\end{corollary}

\section{Summary, future work and conclusion}
Five families of models and consequence relations have been defined and 
their relations will be
summarized here. 
Each family has been characterized by a logical system and no two of those
systems are equivalent.
The family of  cumulative models contains all other
families and is characterized by the logical system {\bf C} that consists
of {\bf Logical Left Equivalence}, {\bf Right Weakening}, {\bf Reflexivity},
{\bf Cut} and {\bf Cautious Monotonicity}.
The next largest family is that of cumulative ordered models. 
It contains all three families not yet mentioned here. It is characterized
by the logical system {\bf CT} that contains, in addition to the rules of 
{\bf C}, the rule of {\bf Loop}.
The families of simple cumulative models and of preferential models are two 
incomparable subfamilies of the family of cumulative ordered models.
Simple cumulative models are characterized by the logical system {\bf CM}
that contains, in addition to the rules of {\bf C}, the rule of {\bf Monotonicity}
(or equivalently, {\bf Transitivity}).
The family of preferential models, probably the most important one,
is characterized by the logical system {\bf P} that contains, in addition
to the rules of {\bf C} the rule of {\bf Or}.
The family of monotonic models is the smallest one of them all. It is contained
in all other four. It is characterized by the logical system {\bf M} that 
contains, in addition to the rules of {\bf C}, both rules {\bf Monotonicity}
and {\bf Or}.

Of those families of consequence relations, which is the best suited to
represent the inferences of a nonmonotonic reasoner in the presence of
a fixed knowledge base? Monotonic and and cumulative monotonic
reasoning are too powerful, i.e. simple cumulative and simple 
preferential models
are too restrictive to represent the wealth of nonmonotonic inference
procedures we would like to consider. We feel that all bona fide logical 
systems should implement reasoning patterns that fall inside the 
framework of cumulative 
reasoning, but probably not all cumulative models represent useful
nonmonotonic systems. The same may probably said about cumulative ordered
models. Preferential reasoning seems to be closest to what we are looking
for.

Nevertheless, many preferential reasoners lack properties that seem 
desirable, for example {\bf Rational Monotonicity}.
A major problem that is not solved in this paper is to describe reasonable
inference procedures that would guarantee that the set of assertions
that may be deduced from any conditional knowledge base satisfies the
property of {\bf Rational Monotonicity}. The second author proposed a solution
to this problem in \cite{Leh:89}.
Another major problem, not solved here, is to extend the results
presented here to predicate calculus and answer the question: how should
quantifiers be treated, or what is the meaning of the conditional
assertion 
\mbox{${\em bird}(x)$ \NI ${\em fly}(x)$}?
The second and third authors have a solution, still unpublished,
to this problem too.

We hope the results presented above will convince the reader that
the field of artificial nonmonotonic reasoning may benefit 
from the study of nonmonotonic consequence relations.
 
\section{Acknowledgments}
Discussions with and remarks from David Makinson, Johan van Benthem,
David Israel,
Benjamin Grosof and an anonymous referee helped us put this work
in perspective and improve its presentation.
They are gratefully acknowledged.


\begin{thebibliography}{10}

\bibitem{Adams:66}
Ernest~W. Adams.
\newblock Probability and the logic of conditional.
\newblock In J.~Hintikka and P.~Suppes, editors, {\em Aspects of Inductive
  Logic}. North Holland, Amsterdam, 1966.

\bibitem{Adams:75}
Ernest~W. Adams.
\newblock {\em The Logic of Conditionals}.
\newblock D. Reidel, Dordrecht, 1975.

\bibitem{Avron:87}
Arnon Avron.
\newblock Simple consequence relations.
\newblock LFCS Report Series 87-30, Dept. of Computer Science, Univ. of
  Edinburgh, June 1987.

\bibitem{Burgess:81}
John~P. Burgess.
\newblock Quick completeness proofs for some logics of conditionals.
\newblock {\em Notre Dame Journal of Formal Logic}, 22:76--84, 1981.

\bibitem{Clark:78}
Keith~L. Clark.
\newblock Negation as failure.
\newblock In H.~Gallaire and J.~Minker, editors, {\em Logics and Data Bases},
  pages 293--322. Plenum Press, 1978.

\bibitem{Del:87}
James~P. Delgrande.
\newblock A first-order logic for prototypical properties.
\newblock {\em Artificial Intelligence}, 33:105--130, 1987.

\bibitem{Del:88}
James~P. Delgrande.
\newblock An approach to default reasoning based on a first-order conditional
  logic: Revised report.
\newblock {\em Artificial Intelligence}, 36:63--90, August 1988.

\bibitem{Eth:85}
David~W. Etherington, Robert~E. Mercer, and Raymond Reiter.
\newblock On the adequacy of predicate circumscription for closed-world
  reasoning.
\newblock {\em Computational Intelligence}, 1:11--15, 1985.

\bibitem{FLM:89}
Michael Freund, Daniel Lehmann, and David Makinson.
\newblock Canonical extensions to the infinite case of finitary nonmonotonic
  inference operations.
\newblock In {\em Workshop on Nomonotonic Reasoning}, pages 133--138, Sankt
  Augustin, FRG, December 1989. Arbeitspapiere der GMD no. 443.

\bibitem{Gabbay:85}
Dov~M. Gabbay.
\newblock Theoretical foundations for non-monotonic reasoning in expert
  systems.
\newblock In Krzysztof~R. Apt, editor, {\em Proc. of the NATO Advanced Study
  Institute on Logics and Models of Concurrent Systems}, pages 439--457, La
  Colle-sur-Loup, France, October 1985. Springer-Verlag.

\bibitem{Gab:pers}
Dov~M. Gabbay.
\newblock personal communication, March 1989.

\bibitem{Gent:32}
Gerhard Gentzen.
\newblock \"{U}ber die existenz unabh\"{a}ngiger axiomensysteme zu unendlichen
  satzsystemen.
\newblock {\em Mathematische Annalen}, 107:329--350, 1932.

\bibitem{Gent:69}
Gerhard Gentzen.
\newblock {\em The Collected Papers of Gerhard Gentzen, edited by M. E. Szabo}.
\newblock North Holland, Amsterdam, 1969.

\bibitem{Gin:86}
Matthew~L. Ginsberg.
\newblock Counterfactuals.
\newblock {\em Artificial Intelligence}, 30:35--79, 1986.

\bibitem{HalpMoses:84}
Joseph~Y. Halpern and Yoram Moses.
\newblock Towards a theory of knowledge and ignorance: preliminary report.
\newblock In {\em Proc. Workshop on Non-Monotonic Reasoning}, pages 125--143.
  AAAI, New Paltz, 1984.

\bibitem{Hoare:69}
Christopher Anthony~R. Hoare.
\newblock An axiomatic basis for computer programming.
\newblock {\em Communications of the ACM}, 12:576--580, 1969.

\bibitem{Isr:80}
David~J. Israel.
\newblock What's wrong with nonmonotonic logic?
\newblock In {\em Proceedings of the AAAI National Conference}, pages 99--101,
  1980.

\bibitem{KLMTR:88}
Sarit Kraus, Daniel Lehmann, and Menachem Magidor.
\newblock Preferential models and cumulative logics.
\newblock Technical Report TR 88-15, Leibniz Center for Computer Science, Dept.
  of Computer Science, Hebrew University, Jerusalem, November 1988.

\bibitem{L:88}
Daniel Lehmann.
\newblock Preferential models and cumulative logics.
\newblock In Ehud Shapiro, editor, {\em Fifth Israeli Symposium on Artificial
  Intelligence, Vision and Pattern Recognition}, pages 365--381, Tel Aviv,
  Israel, December 1988. Information Processing Association of Israel.

\bibitem{Leh:89}
Daniel Lehmann.
\newblock What does a conditional knowledge base entail?
\newblock In Ron Brachman and Hector Levesque, editors, {\em Proceedings of the
  First International Conference on Principles of Knowledge Representation and
  Reasoning}, Toronto, Canada, May 1989. Morgan Kaufmann.

\bibitem{KL:88}
Daniel Lehmann and Sarit Kraus.
\newblock Non monotonic logics: Models and proofs.
\newblock In {\em European Workshop on Logical Methods in Artificial
  Intelligence}, pages 58--64, Roscoff (Finist\`{e}re) France, June 1988.

\bibitem{LMTR:88}
Daniel Lehmann and Menachem Magidor.
\newblock Rational logics and their models: a study in cumulative logic.
\newblock Technical Report TR 88-16, Leibniz Center for Computer Science, Dept.
  of Computer Science, Hebrew University, Jerusalem, November 1988.

\bibitem{Lewis:71}
David~K. Lewis.
\newblock Completeness and decidability of three logics of counterfactual
  conditionals.
\newblock {\em Theoria}, 37:74--85, 1971.

\bibitem{Lewis:73}
David~K. Lewis.
\newblock {\em Counterfactuals}.
\newblock Harvard University Press, 1973.

\bibitem{Lewis:74}
David~K. Lewis.
\newblock Intensional logics without iterative axioms.
\newblock {\em Journal of Philosophical Logic}, 3:457--466, 1974.

\bibitem{Lif:86}
Vladimir Lifschitz.
\newblock On the satisfiability of circumscription.
\newblock {\em Artificial Intelligence}, 28:17--27, 1986.
\newblock Research Note.

\bibitem{Mak:88}
David Makinson.
\newblock personal communication, April 1988.

\bibitem{Mak:89}
David Makinson.
\newblock General theory of cumulative inference.
\newblock In M.~Reinfrank, J.~de~Kleer, M.~L. Ginsberg, and E.~Sandewall,
  editors, {\em Proceedings of the Second International Workshop on
  Non-Monotonic Reasoning}, pages 1--18, Grassau, Germany, June 1988. Springer
  Verlag.
\newblock Volume 346, Lecture Notes in Artificial Intelligence.

\bibitem{McCarthy:80}
John McCarthy.
\newblock Circumscription, a form of non monotonic reasoning.
\newblock {\em Artificial Intelligence}, 13:27--39, 1980.

\bibitem{McDer:80}
Drew McDermott and Jon Doyle.
\newblock Non-monotonic logic {I}.
\newblock {\em Artificial Intelligence}, 13:41--72, 1980.

\bibitem{Moo:84}
Robert~C. Moore.
\newblock Possible-world semantics for autoepistemic logic.
\newblock In {\em Proceedings of AAAI Workshop on Nomonotonic Reasoning}, pages
  396--401, New Paltz, 1984.

\bibitem{Nute:84}
Donald Nute.
\newblock Conditional logic.
\newblock In Dov~M. Gabbay and Franz Guenthner, editors, {\em Handbook of
  Philosophical Logic}, chapter Chapter II.8, pages 387--439. D. Reidel,
  Dordrecht, 1984.

\bibitem{Pearl:88}
Judea Pearl.
\newblock {\em Probabilistic Reasoning in Intelligent Systems: Networks of
  Plausible Inference}.
\newblock Morgan Kaufmann, P.O. Box 50490, Palo Alto, CA 94303, 1988.

\bibitem{PearlGeff:88}
Judea Pearl and Hector Geffner.
\newblock Probabilistic semantics for a subset of default reasoning.
\newblock TR CSD-8700XX, R-93-III, Computer Science Dept., UCLA, March 1988.

\bibitem{Reiter:80}
Raymond Reiter.
\newblock A logic for default reasoning.
\newblock {\em Artificial Intelligence}, 13:81--132, 1980.

\bibitem{Reiter:87}
Raymond Reiter.
\newblock {\em Nonmonotonic Reasoning}, volume~2 of {\em Annual Reviews in
  Computer Science}, pages 147--186.
\newblock Annual Reviews Inc., 1987.

\bibitem{Scott:71}
Dana~S. Scott.
\newblock Completeness and axiomatizability.
\newblock In L.~Henkin~\& al., editor, {\em Proceedings of the Tarski
  Symposium, Proc. of Symposia in Pure Mathematics, Vol. 25}, pages 411--435,
  Providence, R.I., June 1971. Association for Symbolic Logic, American
  Mathematical Society.

\bibitem{Shoham:87}
Yoav Shoham.
\newblock A semantical approach to nonmonotonic logics.
\newblock In {\em Proc. Logics in Computer Science}, pages 275--279, Ithaca,
  N.Y., 1987.

\bibitem{Shoham:88}
Yoav Shoham.
\newblock {\em Reasoning about Change}.
\newblock The MIT Press, 1988.

\bibitem{Stal:68}
Robert~C. Stalnaker.
\newblock {\em A Theory of Conditionals}, volume~2 of {\em American
  Philosophical Quarterly Monograph Series (Nicholas Rescher, ed.}, pages
  98--112.
\newblock Blackwell, Oxford, 1968.

\bibitem{Tar:30a}
Alfred Tarski.
\newblock Fundamentale begriffe der methodologie der deduktiven
  wissenschaften.i.
\newblock {\em Monatshefte f\"{u}r Mathematik und Physik}, 37:361--404, 1930.

\bibitem{Tar:30}
Alfred Tarski.
\newblock \"{U}ber einige fundamentale begriffe der metamatematik.
\newblock {\em Comptes Rendus des s\'{e}ances de la Soci\'{e}t\'{e} des
  Sciences et des Lettres de Varsovie}, 23:22--29, 1930.

\bibitem{Tar:35}
Alfred Tarski.
\newblock Grunz\"{u}ge des systemenkalk\"{u}l, erster teil.
\newblock {\em Fundamenta Mathematic\ae}, 25:503--526, 1935.

\bibitem{Tar:56}
Alfred Tarski.
\newblock {\em Logic, Semantics, Metamathematics. Papers from 1923--1938}.
\newblock Clarendon Press, Oxford, 1956.

\bibitem{Tour:86}
David~S. Touretzky.
\newblock {\em The Mathematics of Inheritance Systems}.
\newblock Research Notes in Artificial Intelligence. Pitman, London --- Morgan
  Kaufmann, Los Altos, 1986.

\bibitem{vBent:84}
Johan van Benthem.
\newblock Foundations of conditional logic.
\newblock {\em Journal of Philosophical Logic}, 13:303--349, 1984.

\bibitem{Velt:86}
Frank Veltman.
\newblock {\em Logics for Conditionals}.
\newblock PhD thesis, Filosofisch Instituut, Universiteit van Amsterdam, 1986.

\end{thebibliography}
\end{document}